\documentclass[a4paper,fleqn]{cas-dc}

\usepackage[authoryear]{natbib}
\usepackage{algorithm}
\usepackage{algorithmic}
\usepackage{amsmath}
\usepackage{xcolor}
\definecolor{lightgray-1}{rgb}{0.6, 0.6, 0.6}
\definecolor{lightgray}{HTML}{f0f0f4}
\usepackage{placeins}

\begin{document}
\let\WriteBookmarks\relax
\def\floatpagepagefraction{1}
\def\textpagefraction{.001}
\shorttitle{Leveraging social media news}
\shortauthors{J.K. Krishnan et~al.}

\title [mode = title]{$\mathrm{Hi}^2$-GSLoc: Dual-Hierarchical Gaussian-Specific Visual Relocalization for Remote Sensing}
\tnotemark[1]

\tnotetext[1]{This work was supported by the National Natural Science Foundation of China (NSFC) under Grant No. 42130112 and the Postdoctoral Fellowship Program of CPSF under Grant No. GZB20240986.}
\author[1,2]{Boni Hu}
\credit{Writing - original draft, Writing - review \& editing, Methodology, Investigation, Visualization}
\author[1,2]{Zhenyu Xia}
\credit{Methodology, Dataset collection}
\author[1,2]{Lin Chen}
\credit{Results visualization}
\author[1,2]{Pengcheng Han}
\credit{Dataset collection, Writing– review \& editing}
\author[1,2]{Shuhui Bu}
\cormark[1]
\credit{Conceptualization of this study, Writing– review \& editing}
\affiliation[1]{organization={School of Aeronautics, Northwestern Polytechnical University},
                city={Xi'an},
                postcode={710072}, 
                country={China}}

\affiliation[2]{organization={National Key Laboratory of Aircraft Configuration
Design},
                city={Xi'an},
                postcode={710072}, 
                country={China}}

\cortext[cor1]{Corresponding author E-mail address: bushuhui@nwpu.edu.cn}

\begin{abstract}
Visual relocalization, which estimates the 6-degree-of-freedom (6-DoF) camera pose from query images, is fundamental to remote sensing and UAV applications. Existing methods face inherent trade-offs: image-based retrieval and pose regression approaches lack precision, while structure-based methods that register queries to Structure-from-Motion (SfM) models suffer from computational complexity and limited scalability. These challenges are particularly pronounced in remote sensing scenarios due to large-scale scenes, high altitude variations, and domain gaps of existing visual priors. To overcome these limitations, we leverage 3D Gaussian Splatting (3DGS) as a novel scene representation that compactly encodes both 3D geometry and appearance. We introduce $\mathrm{Hi}^2$-GSLoc, a dual-hierarchical relocalization framework that follows a sparse-to-dense and coarse-to-fine paradigm, fully exploiting the rich semantic information and geometric constraints inherent in Gaussian primitives. To handle large-scale remote sensing scenarios, we incorporate partitioned Gaussian training, GPU-accelerated parallel matching, and dynamic memory management strategies. Our approach consists of two stages: (1) a sparse stage featuring a Gaussian-specific consistent render-aware sampling strategy and landmark-guided detector for robust and accurate initial pose estimation, and (2) a dense stage that iteratively refines poses through coarse-to-fine dense rasterization matching while incorporating reliability verification. Through comprehensive evaluation on simulation data, public datasets, and real flight experiments, we demonstrate that our method delivers competitive localization accuracy, recall rate, and computational efficiency while effectively filtering unreliable pose estimates. The results confirm the effectiveness of our approach for practical remote sensing applications.

\end{abstract}




\maketitle

\begin{figure}
	\centering
	\includegraphics[width=.9\columnwidth]{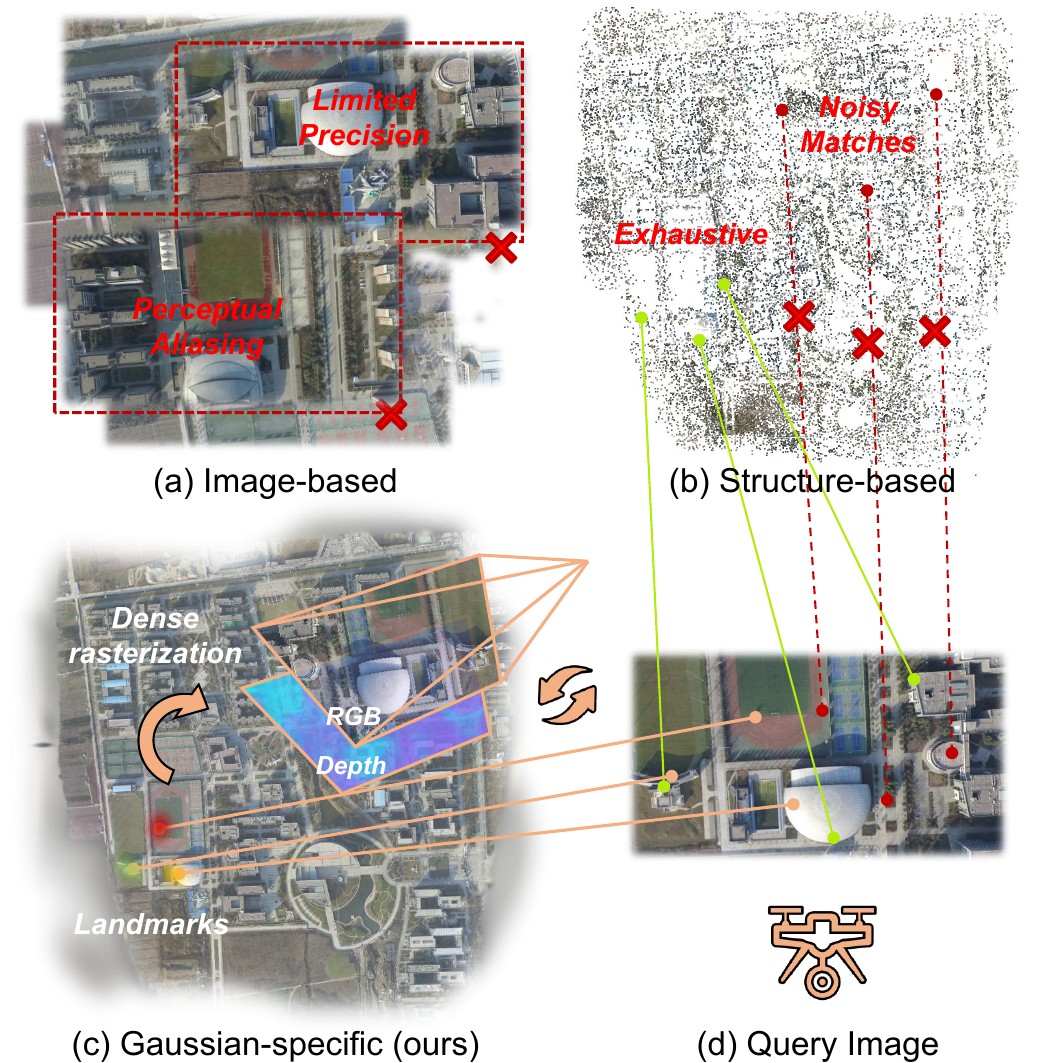}
	\caption{\textbf{Different approaches for visual relocalization.} (a) Image-based: returns location tags from database through image retrieval, or directly regresses 6-DoF pose from images. (b) Structure-based: establishes correspondences between 2D pixels in query images and 3D sparse SfM models, then solves perspective projection optimization equations. (c) Our proposed Gaussian-Specific approach from Sparse Landmarks Sampling to Dense Rasterization Matching.}
	\label{fig:teaser}
\end{figure}

\section{Introduction}

In our increasingly automated world, unmanned aerial vehicles have become indispensable for diverse remote sensing applications—from agricultural monitoring to disaster response and urban planning \cite{isprs-wang2025vecmaplocnet, isprs-ye2024coarse, yin2025general}. At the core of autonomous navigation lies visual relocalization: the ability to estimate precise 6-DoF camera poses from single images against pre-built scene representations. While this capability has been extensively studied for ground-level scenarios, remote sensing environments present unique and formidable challenges that render existing approaches inadequate.

Remote sensing relocalization faces several critical challenges that distinguish it from conventional scenarios. First, the scale disparity is enormous—scenes span kilometers with highly repetitive patterns and sparse distinctive landmarks, making traditional feature matching computationally prohibitive and prone to ambiguous correspondences \cite{isprs-ye2024coarse}. Second, altitude-induced geometric ambiguity creates significant localization uncertainty, as small angular errors propagate to large positional deviations at operational altitudes. Third, existing visual features suffer from severe domain gaps, as most are trained on ground-level imagery and fail to capture the distinctive geometric and photometric characteristics of aerial perspectives. Finally, dramatic viewpoint variations between mapping and query phases, coupled with illumination changes and seasonal variations, can cause catastrophic failures in structure-based methods.

Current relocalization approaches struggle to address these challenges effectively, as shown in Figure \ref{fig:teaser}. Image retrieval methods \cite{berton2025megaloc, keetha2023anyloc, arandjelovic2016netvlad, hu2024curriculumloc}, while demonstrating robustness through contrastive learning, are fundamentally constrained by database density and suffer from perceptual aliasing in repetitive aerial scenes. Direct pose regression approaches \cite{kendall2015posenet, lstm_2016, direct_posenet_2021} lack the geometric grounding necessary for high-precision localization and struggle with generalization across varying scales. Structure-based methods \cite{Sarlin_hloc_2019, brachmann2021visual} achieve superior accuracy through explicit 2D-3D correspondences but face computational bottlenecks in large-scale scenarios and correspondence failures under substantial viewpoint changes. Recent NeRF-based approaches \cite{zhao2024pnerfloc, yen2021inerf} offer promising analysis-by-synthesis capabilities but suffer from prohibitive computational costs and limited real-time applicability.

3DGS presents a compelling solution to these challenges. Unlike implicit neural representations, 3DGS provides explicit, interpretable 3D geometry while maintaining efficient rendering capabilities \cite{Kerbl_gsplat_2023}. Crucially, it encodes both geometric constraints and scene-specific appearance features without relying on external visual priors that may suffer from domain gaps. This makes it particularly well-suited for remote sensing scenarios where traditional visual features often fail. However, existing 3DGS-based relocalization methods \cite{Sidorov_gsplatloc_2025, zhai2025splatloc} are designed for small-scale indoor scenes and lack the specialized components needed for large-scale remote sensing applications. 

To bridge this gap, we introduce $\mathrm{Hi}^2$-GSLoc, a dual-hierarchical relocalization framework specifically tailored for remote sensing scenarios. Specifically, we first estimate an initial pose by registering the input query image to a 3D Gaussian sparse model based on consistent render-aware Gaussian landmark sampling and landmark-guided keypoint detection. Subsequently, we render dense Gaussian rasterization outputs (including feature maps, RGB, and depth) based on the initial pose, then employ coarse-to-fine windowed probabilistic mutual matching with effective iterative refinement to optimize the pose. Finally, consistency-based pose validation filters outliers, achieving accurate and robust relocalization in large-scale remote sensing scenes. Our main contributions are:
\begin{enumerate}[(1)] 
\item We introduce the first 3DGS-based relocalization framework specifically designed for remote sensing scenarios. Our $\mathrm{Hi}^2$-GSLoc employs a dual-hierarchical sparse-to-dense and coarse-to-fine pipeline that integrates partitioned Gaussian training, GPU-accelerated parallel matching, and dynamic memory management strategies to efficiently handle large-scale remote sensing scenes.

\item To address feature domain adaptation and depth ambiguity inherent in remote sensing scenarios, we propose a consistent render-aware landmark sampling strategy (C.R-A.S) coupled with a landmark-guided keypoint detector (L-G.D) that fully exploits geometric constraints and scene-specific representations embedded in Gaussian primitives, enabling robust and accurate initial pose estimation.

\item We design an iterative dense refinement stage that matches rendered Gaussian features with query features through coarse-to-fine windowed probabilistic mutual matching (PMM), coupled with a consistency-based pose validation mechanism to filter unreliable estimates.

\item Extensive experiments validate competitive localization accuracy and recall rates with enhanced robustness through reliable pose filtering, while maintaining computational efficiency.
\end{enumerate}

\section{Related works}   
Visual relocalization research encompasses three primary paradigms: image-based relocalization that operates solely on image information through retrieval or direct pose regression, structure-based relocalization that leverages explicit 3D scene geometry from SfM reconstruction, and analysis-by-synthesis approaches that optimize camera poses through rendering and comparison with query images.

\subsection{Image-based relocalization}
Image-based approaches operate exclusively on visual information without relying on explicit 3D scene structure, broadly categorized into retrieval-based localization and regression-based pose estimation methods.

Retrieval-based methods achieve localization through learned global descriptors. NetVLAD \cite{arandjelovic2016netvlad} pioneered this direction by extracting robust global descriptors via contrastive learning for location retrieval. Recent advances have leveraged foundation vision models \cite{keetha2023anyloc, lu2024towards, lu2025selavpr++, wang2025focus}, developed viewpoint-invariant representations \cite{berton2023eigenplaces}, and introduced comprehensive frameworks that integrate diverse methods, training strategies, and datasets \cite{berton2025megaloc}, achieving substantial improvements. However, these approaches fundamentally depend on database image density and distribution, potentially yielding significant localization errors in sparse coverage scenarios. Moreover, most existing models are predominantly trained on ground-level datasets with limited aerial imagery, leading to substantial domain gaps when applied to remote sensing scenarios.

Regression-based methods directly predict 6-DoF camera poses from single images. PoseNet \cite{kendall2015posenet} introduced the first CNN-based framework for end-to-end pose regression. Subsequent improvements have incorporated temporal information \cite{lstm_2016, viloc_2017}, geometric losses and priors \cite{geometric_loss_2017, geometric_prior_2017}, and photometric consistency constraints \cite{direct_posenet_2021} to enhance pose accuracy. Despite outputting complete 6-DoF poses, these methods typically achieve performance comparable only to image retrieval baselines \cite{arandjelovic2016netvlad} and fall short of structure-based approaches \cite{zhao2024pnerfloc} in terms of precision. Moreover, being inherently data-driven, they exhibit significant performance degradation when applied to domains outside their training distribution.

\subsection{Structure-based relocalization}

Structure-based relocalization methods \cite{Camposeco_hybrid_2018, inloc_Torii_2021, Li_Wang_Zhao_Verbeek_Kannala_2020} leverage 3D scene information reconstructed from SfM to establish 2D-3D correspondences between query images and scenes, subsequently employing Perspective-n-Point (PnP) \cite{pnp_2003, Ke_Roumeliotis_2017} solvers for camera pose estimation. While these approaches fully exploit scene geometry to achieve high pose accuracy, they are susceptible to noisy feature matches and computationally expensive for large-scale scenes. Consequently, image retrieval methods \cite{arandjelovic2016netvlad, berton2023eigenplaces} are typically applied as a preprocessing step to coarsely localize the visible scene structure relative to query images \cite{Sarlin_hloc_2019, inloc_Torii_2021}, significantly reducing localization time. Meanwhile, global image features containing semantic information enhance scene understanding and improve system robustness. Subsequently, 3D point features extracted from the scene image database are matched with 2D keypoint features from query images using identical algorithms to establish 2D-3D correspondences \cite{DeTone_SuperPoint_2018, Revaud_r2d2_2019, Dusmanu_d2net_2019, Sarlin_superglue_2020, lightglue, Sun_loftr_2021, jiang2024omniglue}. HLoc \cite{Sarlin_hloc_2019} integrates diverse global retrieval methods \cite{arandjelovic2016netvlad, berton2023eigenplaces, berton2025megaloc}, feature detectors \cite{DeTone_SuperPoint_2018, tyszkiewicz2020disk}, and feature matchers \cite{Sarlin_superglue_2020, lightglue, jiang2024omniglue} to enhance localization performance.

To mitigate outlier effects, recent advances such as DSAC \cite{brachmann2017dsac, brachmann2021visual} employ CNNs to predict scene coordinates and score hypotheses while introducing differentiable RANSAC algorithms. LoFTR \cite{Sun_loftr_2021} and OmniGlue \cite{jiang2024omniglue} adopt detector-free matching and DINOv2 \cite{oquab2023dinov2} vision foundation models, respectively dedicated to improving robustness under weak texture and large viewpoint variations. 
Despite these advances, structure-based methods are vulnerable to localization failures under substantial viewpoint changes and suffer from computational bottlenecks during feature matching, particularly challenging for large-scale scenarios.

\subsection{Analysis-by-Synthesis}
Analysis-by-synthesis methods optimize camera poses by analyzing relationships between synthesized and query images. These approaches function either as pose correspondence modules or standalone relocalization frameworks, effectively addressing matching failures caused by large viewpoint variations \cite{chen2022dfnet, direct_posenet_2021}. iNeRF \cite{yen2021inerf} directly inverts NeRF models by iteratively optimizing photometric differences between rendered and query images to refine camera pose initialization. 
DirectPN \cite{direct_posenet_2021} integrates NeRF to provide photometric consistency supervision for pose regression by minimizing color discrepancies between query images and those rendered from predicted poses. Dfnet \cite{chen2022dfnet} extends this concept by measuring consistency in feature space, demonstrating enhanced localization performance. PNeRFLoc \cite{zhao2024pnerfloc} employs explicit point-based neural representations to leverage geometric constraints and perform 2D-3D feature matching for 6-DoF pose estimation. However, practical applications remain limited due to NeRF's computationally expensive scene training and view synthesis processes, as well as substantial memory requirements for storing descriptors and correspondence graphs from sparse SfM models.

Compared to NeRF, 3DGS \cite{Kerbl_gsplat_2023} employs explicit representations enabling fast, high-quality view synthesis. Recent advances \cite{feng2025flashgs, wang2024adr, mallick2024taming} have demonstrated real-time, high-fidelity rendering of large-scale scenes using 3DGS. The latest analysis-by-synthesis methods \cite{Sidorov_gsplatloc_2025, zhai2025splatloc, cheng2024logs} integrate 3DGS into relocalization pipelines, combining structure-based coarse pose estimation with photometric rendering optimization in unified end-to-end frameworks. However, existing 3DGS-based relocalization methods exhibit significant limitations for remote sensing applications. They either neglect the rich 3D geometric information embedded in Gaussian primitives \cite{liu2025gs} or directly adapt existing 2D image detectors without Gaussian-specific optimization \cite{Sidorov_gsplatloc_2025}, resulting in suboptimal performance in geometry-sensitive aerial scenarios. While \cite{huang2025sparse} introduced a Gaussian scene-specific detector that improved accuracy, it lacks the memory optimization and scalability strategies essential for large-scale remote sensing environments, limiting its practical applicability.
\begin{figure*}
	\centering
	\includegraphics[width=\textwidth]{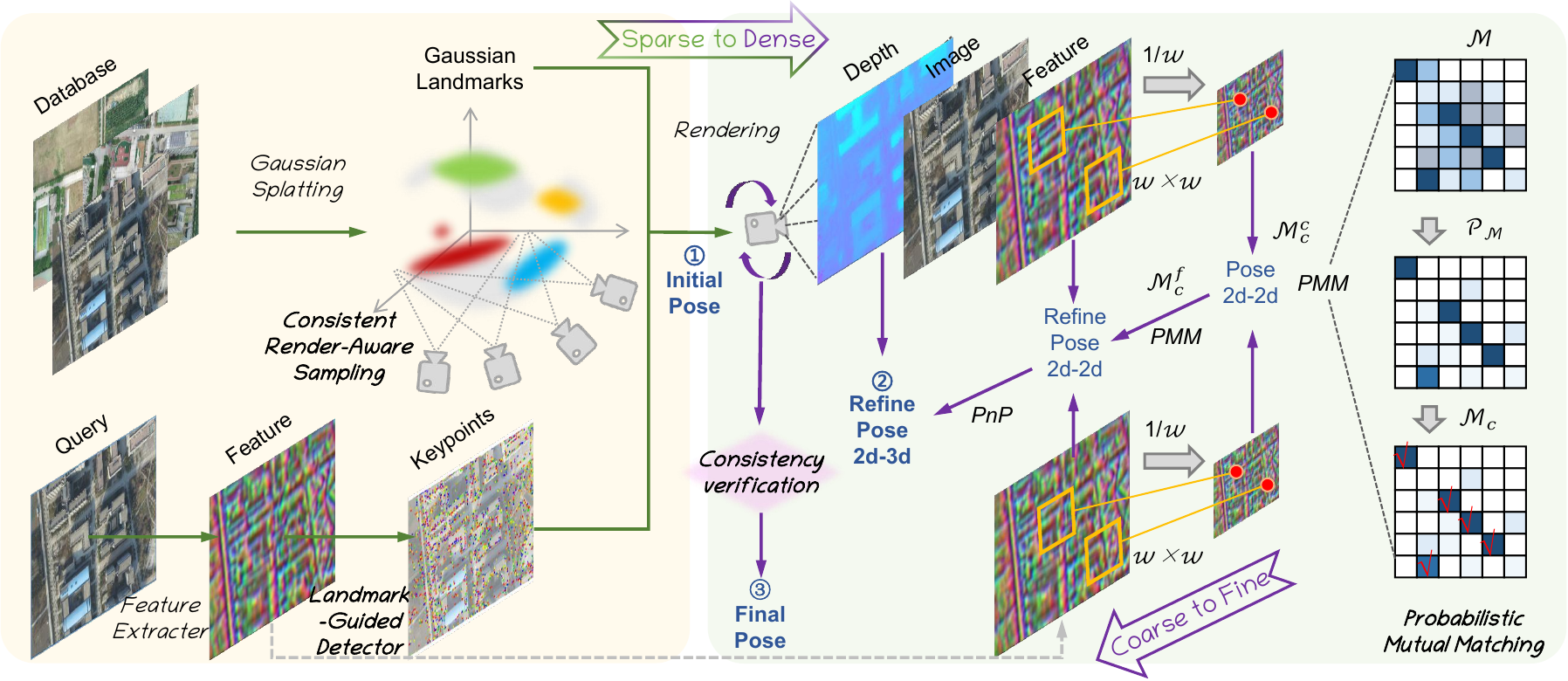}
    \caption{\textbf{Overview of our $\mathbf{\mathrm{Hi}^2}$-GSLoc pipeline.} The method consists of three stages: (1) initial pose estimation through consistent render-aware landmark sampling and landmark-guided keypoint detection, (2) pose optimization via dense rasterization and coarse-to-fine iterative matching, and (3) consistency-based verification to filter unreliable results.}
	\label{fig:outline}
\end{figure*}

\section{Methodlogy}

This section details our dual-hierarchical Gaussian-based relocalization framework, which comprises four core components: (1) 3D Gaussian Splatting foundations and adaptations for remote sensing scenarios, (2) Consistent render-aware Gaussian landmark sampling, (3) Landmark-guided keypoint detection, and (4) Coarse-to-fine dense pose refinement and validation. The complete pipeline is illustrated in Figure \ref{fig:outline}.

\subsection{3D Gaussian splatting for remote sensing}

3DGS \cite{Kerbl_gsplat_2023} represents scenes using millions of 3D Gaussians—colored ellipsoids with transparency that decays according to a Gaussian distribution from their centers. The method initially employs SfM to estimate camera poses and generate sparse point clouds, which are subsequently transformed into initial 3D Gaussians. These Gaussians undergo optimization via Stochastic Gradient Descent (SGD) with adaptive density control, dynamically adding and removing ellipsoids based on gradient magnitudes and predefined criteria to achieve compact, unstructured scene representations. The framework employs tile-based rasterization for efficient real-time rendering of photorealistic scenes.

\begin{figure}
	\centering
	\includegraphics[width=.95\columnwidth]{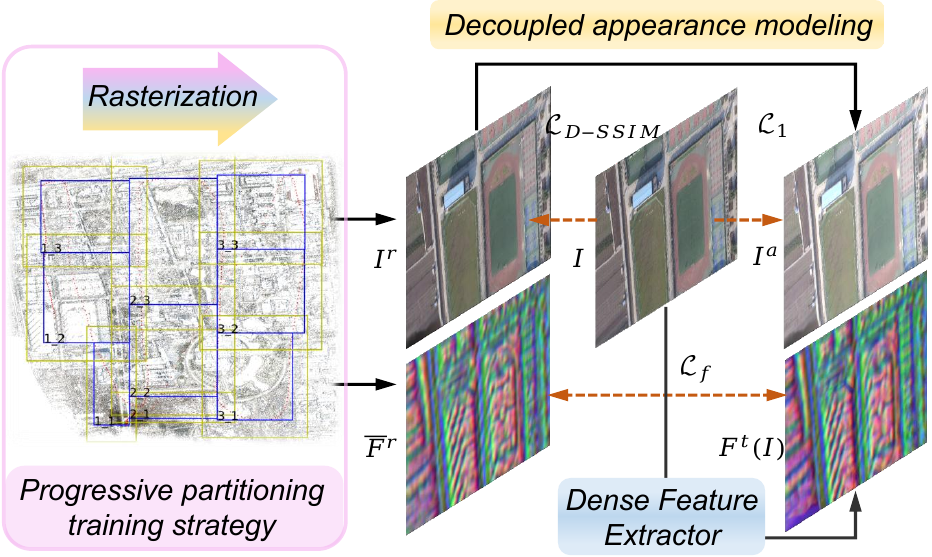}
	\caption{3D feature Gaussian splatting of remote sensing and the training process jointly optimize $\mathcal{L}_{rgb}$ and $\mathcal{L}_{f}$.}
	\label{fig:loss}
\end{figure}

Our approach integrates consistent render-aware sampling strategy and landmark-guided keypoint detector with 3DGS, embedding Gaussian features into 3D representations to enhance relocalization accuracy. Specifically, our scene representation comprises original Gaussian primitives augmented with feature fields.  The trainable attributes of the i-th Gaussian primitive include center $(x_i, y_i, z_i)$, rotation $q_i$, scale $s_i$, opacity $\alpha_i$, color $c_i$, and feature $f_i$, denote as $\Theta_{i}=\{(x_i, y_i, z_i),q_{i},s_{i},\alpha_{i},c_{i},f_{i}\}$. To address challenges in remote sensing large-scale applications including memory constraints, extensive optimization time, and appearance variations, we adopt a partitioning strategy from VastGaussian \cite{lin2024vastgaussianvast3dgaussians}. As shown in the left of Figure \ref{fig:loss}, large scenes are divided into multiple cells using progressive partitioning, where point clouds and training views are allocated to these cells for parallel optimization before seamless merging. Each cell contains fewer 3D Gaussians, enabling optimization within limited memory constraints and reducing training time through parallelization.

The training process follows Feature-3DGS \cite{Zhou_feature_3dgs_2023}, jointly optimizing radiance and feature fields, as illustrated in the right of Figure \ref{fig:loss}. Color attributes $c$ are rasterized into rendered RGB images $I^r$ using alpha blending, while feature attributes $f$ are rendered into feature maps $\bar{F}^r$ through identical rasterization. The ground truth dense feature map extract from the training image $I \in \mathbb{R}^{ 3 \times H \times W}$ is denoted as ${F}^t(I) \in \mathbb{R}^{D \times H' \times W'}$, where $D$ represents the dense feature dimensionality. $F^t(I)$ and query feature maps are both obtained using standard local feature extractors \cite{DeTone_SuperPoint_2018, Revaud_r2d2_2019}. The overall training loss $\mathcal{L}$ combines radiance field loss $\mathcal{L}_{rgb}$ and feature field loss $\mathcal{L}_f$:

\begin{equation}
    \mathcal{L} = \alpha  \mathcal{L}_f + \beta \mathcal{L}_{rgb}.
\end{equation}
The feature field loss $\mathcal{L}_f$ computes the L1 norm loss between ground truth feature maps $F^t(I)$ and rendered feature maps $\bar{F}^r$:

\begin{equation}
    \mathcal{L}_{f}=\frac{1}{N}\sum_{i=1}^{N}|\bar{{F}_i^r}-F_i^t(I)|.
\end{equation}
The radiance field loss $\mathcal{L}_{rgb}$ comprises L1 loss between ground truth images $I$ and appearance-varied rendered images $I^{a}$, and $\mathcal{L}_{D-SSIM}$ loss between directly rendered images $I^{r}$:
\begin{equation}
    \mathcal{L}_{rgb} = (1-\lambda)\frac{1}{N}\sum_{i=1}^{N}|I_i, I_i^a|+\lambda \mathcal{L}_{D-SSIM}(I, I^r),
\end{equation}
where $\mathcal{L}_{D-SSIM}$ denotes the D-SSIM loss \cite{Kerbl_gsplat_2023}, which penalizes structural differences to align structural information in $I^{r}$ with $I$ while L1 loss between appearance-varied rendering $I^{a}$ and $I$ fits ground truth images that may exhibit appearance variations relative to other images. After training, $I^{r}$ achieves consistent appearance across views, enabling 3D Gaussians to learn averaged appearance and correct geometry from all input views. The complete Feature Gaussian scene obtained from this training process is denoted as $\mathcal{G}$.

\subsection{Consistent render-aware sampling}
\begin{figure*}
	\centering
	\includegraphics[width=.9\textwidth]{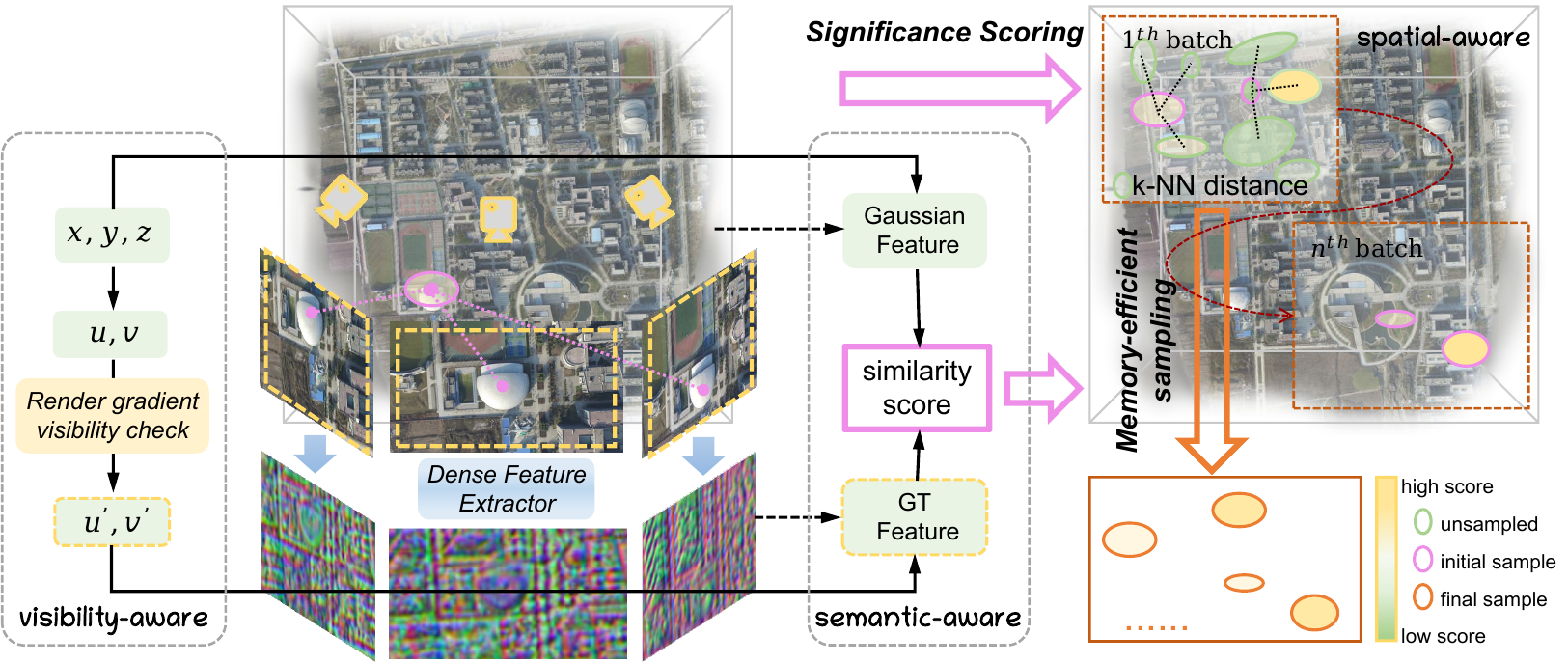}
	\caption{\textbf{Consistent render-aware sampling.} From left to right: significance scoring and memory-efficient sampling strategy based on scores. The entire process incorporates feature and visibility constraints during Gaussian rendering, and spatial distance constraints between Gaussian points.}
	\label{fig:sampling}
\end{figure*}
Exhaustive matching against all Gaussians in a 3DGS model is computationally intensive, and this challenge becomes even more severe in large-scale remote sensing scenarios. Additionally, irrelevant points and Gaussians can easily produce noisy correspondences, degrading localization accuracy. To address this issue, traditional structure-based methods select keypoint-like landmarks through 2D features (e.g., corners, edges, and semantic descriptors). SceneSqueezer \cite{yang2022scenesqueezer} and DetectLandmarks \cite{detectlandmarks2024improved} employ differentiable optimization or learn point importance to reduce map points. SplatLoc \cite{zhai2025splatloc} obtains Gaussian landmarks by learning saliency probability scores of primitives.

In contrast to these landmark selection methods, our approach incorporates visibility, semantic, and geometric constraints throughout the Gaussian rendering process to ensure robust feature matching across different viewpoints. We design batch processing with dynamic memory management to address computational bottlenecks in large-scale scenarios. As illustrated in Figure \ref{fig:sampling}, we first assign saliency scores to each Gaussian primitive by perceiving visibility and semantic features during rendering, then batch-process spatial nearest neighbor groups to select the highest-scoring primitives as landmarks.

\textbf{Significance scoring. \label{sec:score}} Traditional methods are ``feature-driven'': they first detect feature points and then search for matches. Our method is ``geometry-driven'': it establishes correspondences based on 3D geometry and then evaluates feature quality scores. This provides strong geometry priors that ensure correspondences are spatially coherent and physically plausible, reducing the likelihood of outliers and improving overall matching reliability.

The scoring process is illustrated in the left panel of Figure \ref{fig:sampling}. Each camera viewpoint $i$ corresponds to a training image $I$ and a set of visible Gaussian primitives $G_i$. Following a rigorous stereo geometric coordinate transformation pipeline, we compute the transformation of visible Gaussian primitives $G_i$ from world coordinates $(X, Y, Z)$ to camera coordinates and then to pixel coordinates $(U', V') \in I$, thereby obtaining image features $F^t(U',V')$ corresponding to the visible Gaussian primitives features $F_{G_i}$. The matching score $S(G_i)$ of one camera viewpoint $i$ is computed as the cosine similarity between the extracted 2D image features $F^t(U',V')$ at corresponding positions and the Gaussian features $F_{G_i}$:
\begin{equation}
    S(G_i) = \frac{F_{G_i} \cdot {F^t(U',V')}}{||F_{G_i}||_2 \times ||{F^t(U',V')}||_2}. 
\end{equation}
By performing the above operations for each viewpoint, we obtain the visibility count of each Gaussian across different viewpoints, along with the corresponding feature similarity scores. The final similarity score is obtained by averaging across all viewpoints. The total score $S(\mathcal{G})$ of all Gaussian primitives $\mathcal{G}$ across all viewpoints is: 
\begin{equation}
    S(\mathcal{G})=\sum_{i=1}^n S(G_i).
\end{equation}
For the $j^{th}$ Gaussian $g_j$, its final significance score $S(g_j)$ is computed as the average of the total score $S(\mathcal{G_j})$ and visibility count $M$:
\begin{equation}
\label{eq:score-final}
    S(g_j)=\frac{1}{M}S(\mathcal{G}_j) .
\end{equation}
By selecting Gaussian landmarks based on these scores, we can ensure they are easily identifiable and matchable across different viewpoints.

\textbf{Render gradient visibility check. \label{sec:render-check}} Existing methods for matching and landmark selection rely solely on 2D features, ignoring 3D information \cite{leroy2024grounding}. Unlike approaches that determine visibility based only on explicit depth information, we analyze the complete rendering process of Gaussian explicit neural fields. We consider depth occlusion (points occluded by other Gaussian points), opacity (visibility of semi-transparent Gaussian points), rendering weights (actual contribution to the final rendering), and other factors to determine point visibility. This enables us to obtain Gaussian landmarks that are easily identifiable across different viewpoints. Specifically, this is based on gradient determination during the backpropagation process—only Gaussian primitives that contribute to the final rendered image receive gradients during backpropagation. The detailed procedure is shown in Algorithm \ref{algo:render_mask}, where we obtain the final image coordinates $(U',V')$ corresponding to visible Gaussians $(X,Y,Z)$ based on rendering visibility and image projection bounds.

\begin{algorithm}
    \caption{Render gradient visibility check with projection filtering}
    \label{algo:render_mask}
    \textbf{Input:} Gaussian model $\mathcal{G}$, Camera pose $T_{wc}$, Intrinsic $K$,  Image $I \in \mathbb{R}^{ 3 \times H \times W}$\\
    \textbf{Output:} Visible $(U', V')$
    \begin{algorithmic}[1]
    \STATE Extract Gaussian parameters: $(X,Y,Z), \alpha, q, c, s$
    \STATE $RGB \leftarrow rasterization((X,Y,Z), \alpha, q, c, s, K, T_{wc})$
    \STATE $RGB.sum().backward()$ 
        \FOR{$j \in [1, N]$}
            \STATE $M^r[j] \leftarrow (\|\nabla (X,Y,Z)[j]\| > 0)$
        \ENDFOR
    \STATE $(X,Y,Z).grad.zero\_()$
    
    \STATE \textcolor{lightgray-1}{// Project Gaussians to image space}
    \STATE $(X,Y,Z)_{homo} \leftarrow [(X,Y,Z), \mathbf{1}]$ 
    \STATE $(X,Y,Z)_{cam} \leftarrow (T_{WC} \times (X,Y,Z)_{homo}^T)[:3]$ 
    \STATE $depths: d \leftarrow (X,Y,Z)_{cam}[2]$
    \STATE // Perspective division
    \STATE $(X,Y,Z){cam\_homo} \leftarrow (X,Y,Z)_{cam} / d$ 
    \STATE \textcolor{lightgray-1}{// Project to pixel coordinates}
    \STATE $(U,V) \leftarrow (K \times (X,Y,Z)_{cam\_homo})[:2]$
    
    \STATE \textcolor{lightgray-1}{// Boundary and visibility filtering}
    \STATE $M^i \leftarrow ((U,V)[0] \geq 0) \land ((U,V)[0] < W) \land ((U,V)[1] \geq 0) \land ((U,V)[1] < H)$
    \STATE $M \leftarrow M^i \land M^r$ 
    \STATE $(U',V') \leftarrow (U,V)[:, M]$
    
    \RETURN $(U',V')$
    \end{algorithmic}
\end{algorithm}

\textbf{Memory-efficient sampling.} Higher feature similarity scores indicate that the corresponding Gaussian features are more suitable for matching. However, texture-rich regions tend to have higher Gaussian density, so selecting features based solely on scores may lead to insufficient coverage in other regions, particularly problematic in remote sensing images containing large areas of ground and vegetation. To ensure uniform landmark distribution across the entire scene, we employ a two stage selection strategy. We first obtain initial samples $\mathbb{L}_o = \{l_1, l_2, ..., l_Q\}$ through random sampling, where $Q$ is the number of samples. Then, we conduct score-based competition within the spatial K-nearest neighbors (kNN) of each initial sample to derive the final landmarks $\mathbb{L}$. The final selected landmarks are determined by:
\begin{equation}
    L=\{l_i^*\mid l_i^*=\arg\max_{g\in N_k(l_i)}\mathrm{S}(g),\forall l_i\in L_o\},
\end{equation}
where $N_{k}(l_{i})=\{g\in \mathcal{G}:\|g-l_{i}\|\leq r_{i}\}$ is the kNN neighborhood of a initial gaussian sample $l_i$, $r_i$ is the neighborhood search radius, and $S(g)$ is the significance score of each Gaussian $g$ from Eq.\ref{eq:score-final}.

To address memory constraints in large-scale remote sensing scenarios, we partition the Gaussian processing into manageable batches and implement dynamic memory management that actively releases intermediate computation results after each batch, ensuring efficient memory utilization.

\subsection{Landmark-guided detector}

\begin{figure}
    \centering
    \includegraphics[width=\columnwidth]{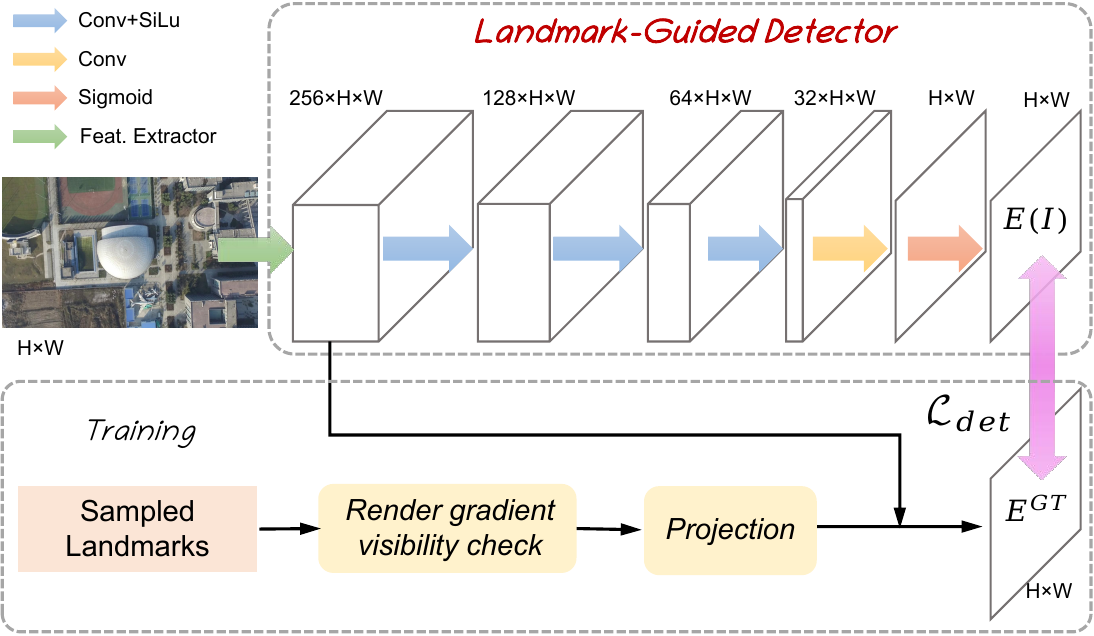}
    \caption{\textbf{Lnadmark-guided detector}. Training 2D image keypoints detector guided by sampled 3D Gaussian landmarks.}
    \label{fig:detector}
\end{figure}

Directly matching dense feature maps with sampled landmarks is infeasible, as dense feature maps contain numerous position-irrelevant and unsuitable redundant features for matching. GSplatLoc \cite{Sidorov_gsplatloc_2025} directly uses cosine similarity between image features extracted by existing 2D image detectors (XFeat \cite{xfeat2024cvpr}) and Gaussian features based on XFeat distillation to obtain matching relationships, without considering 3D information in the Gaussian model or retraining for Gaussian scenes, resulting in significantly reduced accuracy in remote sensing scenarios. Moreover, off-the-shelf detectors \cite{Sun_loftr_2021, DeTone_SuperPoint_2018, Dusmanu_d2net_2019, Revaud_r2d2_2019} typically detect scene-agnostic predefined keypoints, making them unsuitable for matching with sampled landmarks in feature Gaussian scenes. To address this problem, we train a Gaussian-specific landmark-guided detector that can process feature maps $F^t(I)$ and generate a probability map $E(I) \in \mathbb{R}^{1 \times H \times W}$, representing the probability of 2D features being landmarks. The network architecture and training pipeline are illustrated in the Figure \ref{fig:detector}. Specifically, our detector $D_\theta(F^t(I))$ is a shallow CNN appended after existing feature extractors \cite{DeTone_SuperPoint_2018, xfeat2024cvpr}, where $\theta$ represents the network parameters.

The training process is conducted in a self-supervised manner, leveraging 3D geometric constraints to enhance the quality and consistency of 2D feature point detection. Similar to determining robust matching points visible across different viewpoints through render gradient visibility checks, our detector aims to detect Gaussian points that are render-visible in the current image viewpoint. Specifically, we project the center of each Gaussian from the selected Gaussian landmarks onto the current image plane and obtain ground truth Gaussian matching points for the current viewpoint based on rendering visibility. We then use binary cross-entropy loss to optimize the detector $D_\theta$:

\begin{equation}
    \begin{aligned}
        & \mathcal{L}_{det}(E(I),E^{GT}) = -\frac{1}{H \times W}\sum_{h=1}^{H}\sum_{w=1}^{W} \\
        & \quad [E^{GT}_{h,w}\log(E(I)_{h,w}) + (1-E^{GT}_{h,w})\log(1-E(I)_{h,w})]
    \end{aligned}
\end{equation}

During inference, non-maximum suppression (NMS) is applied to the output probability map of the trained detector to ensure uniform distribution of detected keypoints. The final detected keypoints can be represented as follows:
\begin{equation}
    \begin{aligned}
        \mathcal{K} = \{(u_i, v_i) \mid E(u_i, v_i) > \tau \enspace\&\enspace E(u_i, v_i) = \max_{\mathcal{N}_r(u_i, v_i)} E(u,v)\},
    \end{aligned}
\end{equation}
where $\mathcal{K}$ represents the final set of detected keypoints, $(u_i, v_i)$ are the pixel coordinates of keypoints, $\tau$ is the confidence threshold, $r$ is the NMS suppression radius, and $\mathcal{N}_r(u_i, v_i)$ denotes the neighborhood of radius $r$ centered at $(u_i, v_i)$.

The entire pipeline follows rigorous stereo geometric coordinate transformations to generate supervision information, embodying the ``3D-guides-2D'' philosophy. Therefore, the detected keypoints and subsequently established matches possess more accurate geometric relationships, ensuring precise localization even in scale-sensitive scenarios such as remote sensing.

\subsection{Dense rasterization matching}

Using the 3D landmarks obtained from C.R-A.S and the 2D keypoints detected by the L-G.D, we establish the top-$k$ 2D-3D correspondences based on cosine similarity between features and solve PnP to obtain the initial pose:

\begin{equation}
    \label{eq:slove-pnp}
    \{R^*,t^*\},\mathcal{I}^*=\arg\min_{R,t}\sum_{i\in\mathcal{I}}\rho(\|p_i^{2D}-\pi(K[R|t]\tilde{g_i^{3D}})\|_2,\tau),
\end{equation}
where $\rho(\cdot, \tau)$ is the robust loss function with threshold $\tau$ (reprojection error), $\pi(\cdot)$ is the projection function: $\pi([x,y,z]^T) = [x/z, y/z]^T$, $\mathcal{I}^*$ is the optimal inlier set, $K$ is the camera intrinsic matrix, $p^{2D}_i\in \mathbb{R}^2$ is the 2D point on the query image plane $g^{3D}_i \in \mathbb{R}^3$ is the 3D Gaussian point in world coordinates. 
As shown in Figure \ref{fig:outline}, based on the initial pose, we can render dense feature and depth maps from the complete Feature Gaussian scene $\mathcal{G}$, and then iteratively optimize the pose through coarse-to-fine dense feature matching.

\textbf{Coarse to fine pose refinement.} To improve computational efficiency, we first perform matching on coarse query and rendered feature maps, reducing the search space from $O (H_f\times W_f \times H_f \times W_f)$ to $O (H_c \times W_c \times H_c \times W_c)$. In this work, we set $H_f/H_c=8$, thereby achieving a $4096$-fold reduction in search space while avoiding the storage of massive matrices generated by high-resolution image processing in remote sensing. 

Subsequently, we perform sliding window-based sub-pixel level matching on fine query and rendered feature maps. This approach ensures accurate pose estimation precision while employing parallel GPU computation across windows, significantly improving efficiency. The window size is $w = H_f/H_c $, which adaptively accommodates different resolutions. Finally, we fully leverage the depth information from Gaussian rendering for 3D constraints. The specific solution is based on 2D-3D PnP algorithm with RANSAC, as described in Eq. \ref{eq:slove-pnp}. 

\textbf{Consistency verification.} To prevent large errors in initial pose estimation that could cause significant viewpoint differences between dense rasterization and query views, we iteratively execute $n$ rounds of rendering and coarse-to-fine pose optimization with the optimized dense pose, and perform consistency pose filtering by checking angular differences across multiple results. In practice, $n$ is set to 3, as computed in Algorithm. \ref{algo:pose_consistency}, ``$\text{trace} \leftarrow \min(3.0, \max(\text{trace}(R_{rel}), -1.0))$'' is used to prevent computational collapse due to rounding errors. During a single pose computation, if we detect pose inconsistency between any two coarse-to-fine iterative dense pose calculations—i.e., angular difference exceeding threshold $\tau=20^\circ$—we consider the result unreliable and directly skip to compute the next query, ensuring the localization system is not affected by erroneous results under extreme conditions.

\begin{algorithm}
    \caption{Pose Consistency Verification}
    \label{algo:pose_consistency}
    \textbf{Input:} Dense pose results $\{T_i\}_{i=1}^{n}$, Threshold $\tau = 20^\circ$ \\
    \textbf{Output:} Final pose or Status
    \begin{algorithmic}[1]
    \STATE \textcolor{lightgray-1}{ // Compute Pose Difference}
    \STATE \textbf{Function} $\Psi(T_1, T_2)$:
    \STATE \quad $R_1 \leftarrow T_1[:3, :3]$, $R_2 \leftarrow T_2[:3, :3]$
    \STATE \quad $t_1 \leftarrow T_1[:3, 3]$, $t_2 \leftarrow T_2[:3, 3]$ 
    \STATE \quad $R_{rel} \leftarrow R_1 \times R_2^T$ 
    \STATE \quad \textcolor{lightgray-1}{ // Numerical Stabilization}
    \STATE \quad $trace \leftarrow min(3.0, max(\text{trace}(R_{rel}), -1.0))$ 
    \STATE \quad $\theta_{diff} \leftarrow \frac{180}{\pi} \arccos\left(\frac{trace - 1}{2}\right)$
    \STATE \quad $d_{trans} \leftarrow \|t_1 - t_2\|_2$
    \STATE \quad return $\theta_{diff}, d_{trans}$
    \STATE \textbf{EndFunction}
    
    \STATE \textcolor{lightgray-1}{ // Main consistency verification loop}
    \FOR{$i = 1$ \textbf{to} $n-1$}
        \STATE $\theta_{diff}, d_{trans} \leftarrow$ $\Psi(T_1, T_2)$
        \IF{$\theta_{diff} > \tau$}
            \STATE return \ ``unreliable''
        \ENDIF
    \ENDFOR
    \STATE return \ final\_pose: $T_n$
    \end{algorithmic}
\end{algorithm}

\textbf{Probabilistic mutual matching.} During the dense rasterization matching process, relying solely on cosine similarity can easily produce many-to-one and one-to-many mismatches. Therefore, we design probabilistic mutual matching. As shown in Figure \ref{fig:outline}, we compute the cosine similarity between the query feature map and rendered feature map to obtain matrix $\mathcal{M}$, then calculate bidirectional softmax to get the mutually constrained probability matrix $\mathcal{P_M}$:
\begin{equation}
    \mathcal{P_M} = \frac{\exp(\mathcal{M} / \tau)}{\sum_j \exp(\mathcal{M}_{ij} / \tau)} \odot \left(\frac{\exp(\mathcal{M}^T / \tau)}{\sum_i \exp(\mathcal{M}_{ji} / \tau)}\right)^T
\end{equation}
where $\tau$ is the temperature parameter that can adjust the retention of matching relationships with different confidence levels. Finally, we apply mutual nearest neighbor (MNN) search on $\mathcal{P_M}$ to establish correspondences $\mathcal{M}_c$. The entire coarse-to-fine matching process executes the above operations, where fine feature map matching $\mathcal{M}_c^f$ is generated from ${w} \times {w}$ windows extracted at each position of the coarse matching. This operation significantly enhances the quality of dense matching $\mathcal{M}_c^c$ and $\mathcal{M}_c^f$, leading to more accurate pose estimation.

\section{Experiments and analysis}

In this section, we first introduce the benchmark datasets and a remote sensing dataset we collected, then outline the experimental settings and evaluation metrics. Finally, we provide detailed analysis of performance comparisons between our proposed method and other existing approaches, along with ablation studies.

\subsection{Datasets}

To comprehensively evaluate the effectiveness and robustness of our proposed localization methods, we conduct extensive experiments across three categories of datasets representing different scales and deployment scenarios.

\textbf{Standard outdoor localization dataset.} We first evaluate our method against state-of-the-art approaches using the widely adopted Cambridge Landmarks dataset \cite{kendall2015posenet}, which comprises five outdoor scenes captured with mobile phones. This dataset presents typical visual localization challenges including dynamic object occlusion, illumination variations, and motion blur. 

\textbf{Large-scale aerial dataset.} For large-scale scenarios, we utilize the Mill 19-Rubble dataset \cite{turki2022mega}, which provides extensive aerial imagery suitable for evaluating UAV localization algorithms under challenging real-world conditions.

\begin{figure}
    \centering
    \includegraphics[width=\columnwidth]{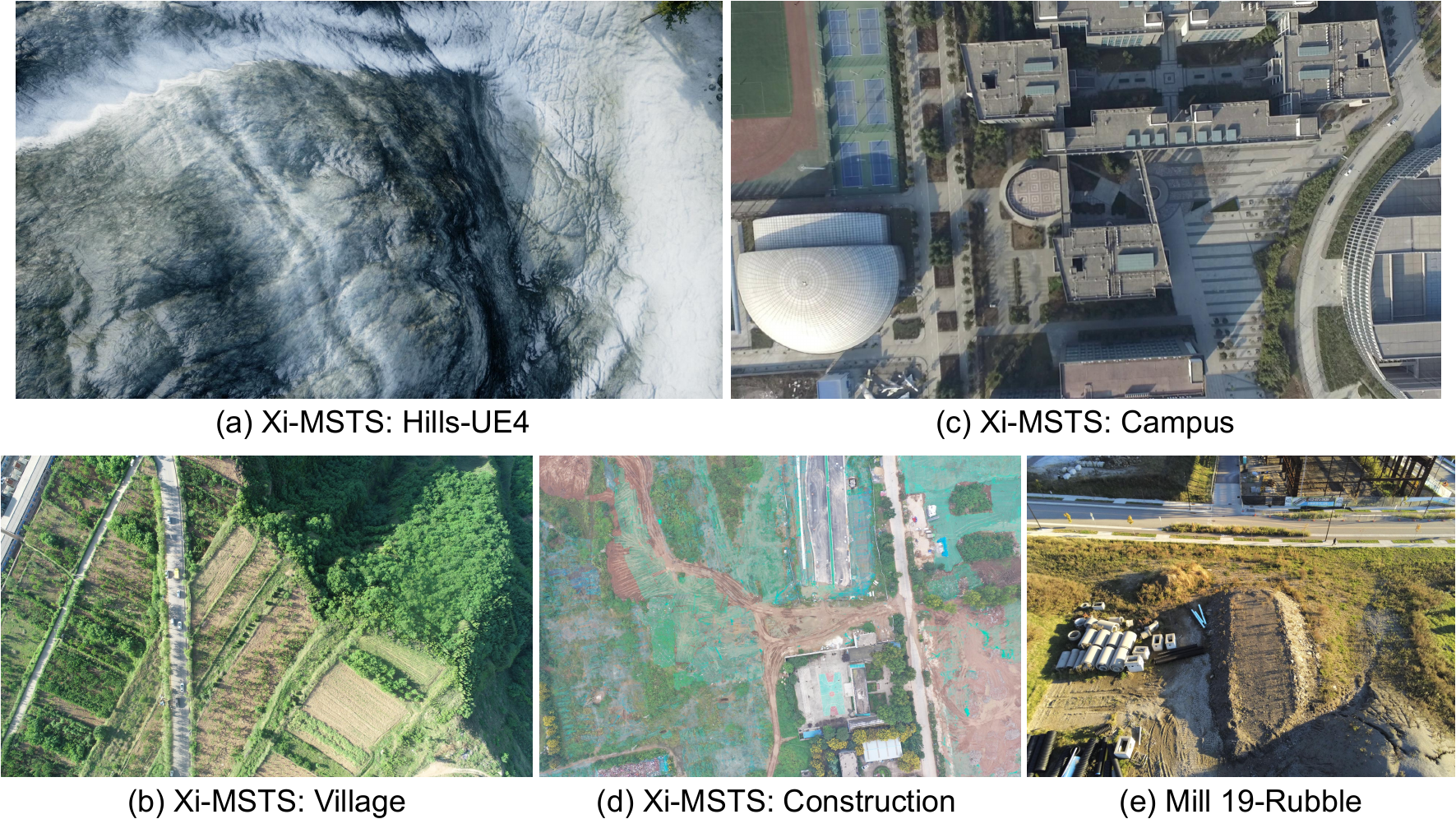}
    \caption{Representative image samples from five diverse scenes in the Xi-MSTS dataset and Mill 19-Rubble, showcasing significant heterogeneity in spatial scales, terrain characteristics, and imaging conditions.}
    \label{fig:dataset}
\end{figure}

\textbf{Xi-MSTS.} To evaluate algorithm performance across varied deployment conditions and validate the effectiveness across different data modalities, we construct the Xi-MSTS (Xi'an Multi-Scene Temporal Sensing) dataset, which comprises both real-world and synthetic scenarios. Specifically, the dataset includes three real-world scenes captured within Xi'an, China: Village, Construction and Campus, spanning multiple years (2016-2020) with significant heterogeneity in spatial scales, terrain characteristics, and imaging conditions. Additionally, we include one synthetic scene (Hills-UE4) generated using Unreal Engine 4 to assess algorithm generalization in different applications. The real-world scenes are captured using different UAV platforms and camera configurations, with high-precision ground truth poses obtained through RTK-GPS measurements, ensuring centimeter-level positioning accuracy. The synthetic scene provides controlled experimental conditions with known ground truth. Figure \ref{fig:dataset} presents representative samples from Xi-MSTS, and Table \ref{tab:dataset} provides detailed statistics and technical specifications. This diverse dataset, encompassing both real and synthetic environments, enables comprehensive evaluation of algorithm robustness and cross-domain generalization capability across various deployment scenarios.

\begin{table*}[!t]
\caption{Dataset summary.}
\label{tab:dataset}
\begin{tabular*}{\tblwidth}{L|LLLLLLL}
\toprule
Datasets & Numbers & Resolution & View & Height/$m$ & Sensor & Area/$km^2$ & Acquisition time\\
\midrule
Cambridge Landmarks & 4991 & $1920 \times 1080$ & Ground-view & None & Phone & 0.013 & 2015 \\
\midrule
Mill 19-Rubble & 1678 &  $4608 \times 3456$  & Oblique photography & Low & None & None& 2022 \\
\midrule
Xi-MSTS: Hills-UE4 & 538 & $1920 \times 1080$ & Terrain following &  None & AirSim &0.2540 & 2025 \\ 
Xi-MSTS: Village   & 709  & $6475 \times 3906$ & Vertical photography & 830-845   &  DJI &  1.0056 &2020-0512-6pm. \\
Xi-MSTS: Campus      & 457  & $1920 \times 1080$ & Vertical photography &   485 & DJI & 1.0735 & 2016-01-26 \\
Xi-MSTS: Construction   & 533  & $5472 \times 3648$ & Vertical photography & 626-647 & Hasselblad & 0.8332 & 2019-09-12-11am.  \\
\bottomrule
\end{tabular*}
\end{table*}

\begin{table*}[!t]
\caption{Quantitative comparison of various advanced methods on the Cambridge Landmarks dataset. The results are shown below, with \textcolor{red}{red} indicating the best performance and \textcolor{blue}{blue} indicating the second best.}
\label{tab:cambridge}
    \begin{tabular*}{\textwidth}{@{}@{\extracolsep{\fill}}l|l|lllll@{}}
        \toprule
        & Methods & Kings & Hospital & Shop & Church & Avg.$\downarrow$ {[}cm/$^ \circ${]} \\ \hline
        \multirow{5}{*}{\begin{tabular}[c]{@{}l@{}}Imaged-\\ based\end{tabular}} & PoseNet  & 166/4.86 & 262/4.90 & 141/7.18 & 245/7.95 & 204/6.23 \\ 
        & MS-Transformer & 83/1.47 & 181/2.39 & 86/3.07 & 162/3.99 & 128/2.73 \\ 
        & Learn-$\theta^2$ PN & 99/1.06 & 217/2.94 & 105/3.97 & 149/3.43 & 143/2.85 \\  
        & LSTM PN  & 99/3.65 & 151/4.29 & 118/7.44 & 152/6.68 & 130/5.51 \\
        & Geo. PN & 88/1.04 & 320/3.29 & 88/3.78 & 157/3.32 & 163/2.86 \\ 
        \midrule
        \multirow{3}{*}{\begin{tabular}[c]{@{}l@{}}Structure-\\ based\end{tabular}} 
        & SIFT  & 13/0.22 & 20/0.36 & \textcolor{blue}{4.0}/0.21 & 8.0/0.25 & 11.25/0.26 \\ 
        & HSCNet & 18/0.30 & 19/0.30 & 6/0.30 & 9.0/0.30 & 13.0/0.30 \\ 
        & HLoc (SP +SG)& \textcolor{red}{11}/\textcolor{blue}{0.20} & \textcolor{blue}{15.1}/\textcolor{blue}{0.31} & 4.2/\textcolor{blue}{0.20} & \textcolor{blue}{7.0}/\textcolor{blue}{0.22} & \textcolor{blue}{9.3}/\textcolor{blue}{0.23} \\  
        & DSAC*  & 17.9/0.31 & 21.1/0.40 & 5.2/0.24 & 15.4/0.51 & 14.9/0.37 \\
        \midrule
        \multirow{6}{*}{\begin{tabular}[c]{@{}l@{}}Analysis\\ -by-\\ synthesis\end{tabular}} & Dfnet  & 43/0.87 & 46/0.87 & 16/0.59 & 50/1.49 & 39/0.96 \\ 
        & NeRFMatch & 12.5/0.23 & 20.9/0.38 & 8.4/0.40 & 10.9/0.35 & 13.2/0.34 \\ 
        & PNeRFLoc & 24/0.29 & 28/0.37 & 6.0/0.27 & 40/0.55 & 24.5/0.37 \\
        & CROSSFIRE & 47/0.7 & 43/0.7 & 20.0/1.2 & 39/1.4 & 37.3/1.00 \\
        & GSplatLoc  & 31/0.49 & 16/0.68 & \textcolor{blue}{4.0}/0.34 & 14/0.42 & 16.25/0.49 \\ 
        & $\mathrm{Hi}^2$-GSLoc & 14.6/\textcolor{red}{0.15} &\textcolor{red}{11.5}/\textcolor{red}{0.21} & \textcolor{red}{2.9}/\textcolor{red}{0.12} & \textcolor{red}{4.6}/\textcolor{red}{0.13} & \textcolor{red}{8.4}/\textcolor{red}{0.15} \\ \bottomrule
    \end{tabular*}
\end{table*}

\subsection{Implementation details}

Our training configuration follows VastGaussian \cite{lin2024vastgaussianvast3dgaussians} with modifications for feature learning. The feature field is trained with a learning rate of 0.001, following Feature 3DGS \cite{Zhou_feature_3dgs_2023}. All scenes undergo training for 30,000 iterations. The densification process is scheduled from iteration 500 to 20,000 with an interval of 500 iterations. This progressive densification allows adaptive scene representation refinement while maintaining training stability. To manage computational complexity while preserving essential details, we employ different resolution settings: Xi-MSTS-Village is trained at 1/4 resolution, while other Xi-MSTS scenes, Cambrideg Landmarks and Mill 19-Rubble are trained at 1/2 resolution. For sparse matching, we extract 16,384 landmarks per scene to ensure sufficient spatial coverage for relocalization. And the Landmark-Guided keypoint detector is trained for 30,000 iterations using a learning rate of 0.001 with cosine decay scheduling. This learning rate schedule ensures stable convergence while preventing overfitting to specific scenes. All experiments are conducted on a single RTX 3090 GPU. Training times are approximately 150 minutes or less for Feature Gaussian optimization and under 50 minutes for scene-specific detector training per scene, demonstrating the practical efficiency of our approach.

\subsection{Evaluation metric}

We employ two complementary metrics to comprehensively evaluate localization performance. The median localization error quantifies both translational and rotational accuracy: translational error (TE) measures the Euclidean distance between ground truth and estimated camera positions, while angular error (AE) captures the angular deviation between ground truth and predicted camera orientations. The localization recall rate represents the percentage of test images successfully localized within predefined error thresholds. Specifically, an image is considered successfully localized when both translational and rotational errors fall below specified tolerance levels simultaneously. These metrics collectively provide a comprehensive assessment of typical accuracy (via median error) and overall system reliability (via recall rate) for each evaluated method.
\begin{table*}[!t]
\caption{Quantitative comparison of state-of-the-art methods on various kind of remote sensing dataset with SfM ground truth. The results are shown below, with red and blue indicating the best and second-best performance across our unfiltered estimates and other methods.}
\label{tab:compare}
    \begin{tabular*}{\textwidth}{l|l|ll|llll|l}
    \toprule
     & method & AE $\downarrow$ & TE $\downarrow$ & 500/10$^ \circ$ $\uparrow$ & 200/5$^ \circ$ $\uparrow$& 5/5$^ \circ$ $\uparrow$& 2/2$^ \circ$ $\uparrow$ & Inference/s $\downarrow$ \\ \midrule
     & SP+SG & 4.4167 & 66.9698 & 50.3 & 50.25 & 1.01 & 0.00 & 11.9648\\ 
     & disk+LG & 2.1353 & 49.1413 & 50.3 &  50.25 & 0.00 & 0.00 & 33.2160 \\ 
     & NetVLAD+disk+LG &  1.9572 &  30.237 & 90.00 &  90.00 &  0.00 & 0.00 & \textcolor{blue}{4.8623}\\ 
     & MegaLoc+disk+SG & \textcolor{blue}{1.9385} &  \textcolor{blue}{28.4464} &\textcolor{red}{100.0} &  \textcolor{red}{100.0} & 0.00 & 0.00 & 4.9951 \\ 
     & Eigenplaces+disk+LG & 2.1647 & 49.1749 & 50.3 & 50.25 & 0.00 & 0.00 & 4.8997 \\ 
     & GSplatLoc & 108.34 & 959.34 & 5.23 & 0.65 & 0.00 & 0.00 & {8.232} \\ 
     & ours & \textcolor{red}{0.0128} & \textcolor{red}{0.1021} & \textcolor{blue}{93.87} & \textcolor{blue}{93.87} & \textcolor{red}{93.87} & \textcolor{red}{93.87} & \textcolor{red}{2.4689} \\ 
     \multirow{-9}{*}{Mill 19-Rubble} &\cellcolor{lightgray}ours(final) &\cellcolor{lightgray}{0.0119} &\cellcolor{lightgray}{0.0997} &\cellcolor{lightgray}{100.0} &\cellcolor{lightgray}{100.0} &\cellcolor{lightgray}{100.0} &\cellcolor{lightgray}{100.0} &\cellcolor{lightgray}0.00024\\ 
     \midrule 
     & SP+SG & 1.2779 & 30.7426 & 58.4 & 57.83 & 0.00 & 0.00 & 10.1121 \\  
     & disk+LG &  8.8689 & 89.0069 & 55.1 & 0.00 & 0.00 & 0.00 & 22.5649 \\ 
     & NetVLAD+disk+LG & 7.4086 & 69.4838 & 56.2 & 0.00 & 0.00 & 0.00 & 3.3210 \\  
     & MegaLoc+disk+SG & \textcolor{blue}{4.1653} & \textcolor{blue}{33.5057} & \textcolor{blue}{90.7} & \textcolor{blue}{89.71} & 0.00 & 0.00 & 3.4698 \\ 
     & Eigenplaces+disk+LG & 8.2131 & 95.8683 & 47.6 & 0.00 & 0.00 & 0.00 & \textcolor{blue}{3.0021} \\  
     & GSplatLoc & 95.434 & 453.334 & 8.5 & 6.5 & \textcolor{blue}{3.3} & \textcolor{blue}{2.6} & {8.213} \\ 
     & ours & \textcolor{red}{0.1062} & \textcolor{red}{0.4552} & \textcolor{red}{98.19} & \textcolor{red}{98.19} & \textcolor{red}{98.19} &\textcolor{red}{98.19} & \textcolor{red}{1.1209} \\  
     \multirow{-9}{*}{Hills-UE4} &\cellcolor{lightgray}ours(final) &\cellcolor{lightgray}{0.1050}   &\cellcolor{lightgray}{0.4574}  & \cellcolor{lightgray}{100.0} &\cellcolor{lightgray}{100.0}  &\cellcolor{lightgray}{100.0}   &\cellcolor{lightgray}{100.0} &\cellcolor{lightgray}0.00025 \\ \midrule
     & SP+SG & 3.1137 & 8.9276 & \textcolor{red}{100.0} & \textcolor{red}{100.0} & 19.66 & 0.00 &13.7490\\ 
     & disk+LG & 1.2339 & 3.8313 & \textcolor{red}{100.0}& \textcolor{red}{100.0} & \textcolor{blue}{60.11} & 26.96 & 38.5623\\ 
     & NetVLAD+disk+LG & \textcolor{blue}{1.1983} & \textcolor{blue}{3.5315} & \textcolor{red}{100.0} & \textcolor{red}{100.0} & 59.55 & 29.21 & \textcolor{blue}{5.4126}\\ 
     & MegaLoc+disk+SG & 1.2304 & 3.6837 & \textcolor{red}{100.0} & \textcolor{red}{100.0} & 58.43 & \textcolor{blue}{33.14} & 5.6213\\ 
     & Eigenplaces+disk+LG & 1.2581 & 3.7340 & \textcolor{red}{100.0} & \textcolor{red}{100.0} & 58.98 & 30.89 & 5.6379\\ 
     & GSplatLoc & 3.2028 & 313.9821 & 75.42 & 20.00 & 2.85 & 2.85 & {8.562} \\ 
     & ours & \textcolor{red}{0.0291} & \textcolor{red}{0.1456} & \textcolor{red}{100.0} & \textcolor{red}{100.0} & \textcolor{red}{100.0} & \textcolor{red}{100.0} & \textcolor{red}{5.0126} \\ 
     \multirow{-9}{*}{Construction}  &\cellcolor{lightgray}ours(final) &\cellcolor{lightgray}{0.0291} &\cellcolor{lightgray}{0.1456} &\cellcolor{lightgray}{100.0} &\cellcolor{lightgray}{100.0} &\cellcolor{lightgray}{100.0} &\cellcolor{lightgray}{100.0} & \cellcolor{lightgray}0.00028\\ \midrule
     & SP+SG & \textcolor{blue}{3.2101} & \textcolor{blue}{19.8647} & \textcolor{blue}{100.0} & \textcolor{blue}{95.63} & 0.00 & 0.00 & 10.0601  \\ 
     & disk+LG & 10.0266 & 41.5737 & 49.8 & 0.00 & 0.00 &0.00 & 21.3221 \\ 
     & NetVLAD+disk+LG & 10.0365 & 41.3977 & 49.3 & 0.00 &0.00  & 0.00 & 3.0146 \\ 
     & MegaLoc+disk+SG &  10.0564& 41.0556 & 47.6 & 0.00 & 0.00 & 0.00 & 3.2234 \\ 
     & Eigenplaces+disk+LG &  10.0737& 41.5415 & 48.0 & 0.00 & 0.00 & 0.00 & \textcolor{blue}{3.0126}\\ 
     & GSplatLoc & 73.8292 & 447.6325 & 31.37 & 26.14 & 9.81 & 1.96 & {8.2341} \\ 
     & ours & \textcolor{red}{0.0377} & \textcolor{red}{0.1578} & \textcolor{red}{98.69} & \textcolor{red}{98.69} & \textcolor{red}{98.69} & \textcolor{red}{98.69} & \textcolor{red}{1.5514} \\
     \multirow{-9}{*}{Campus} &\cellcolor{lightgray}ours(final) &\cellcolor{lightgray}0.0362 &\cellcolor{lightgray}0.1552 &\cellcolor{lightgray}100.0 &\cellcolor{lightgray}100.0 &\cellcolor{lightgray}100.0 &\cellcolor{lightgray}100.0 &\cellcolor{lightgray}0.00026 \\ 
     \bottomrule
    \end{tabular*}
\end{table*}

\begin{figure*}
	\centering
	\includegraphics[width=.95\textwidth]{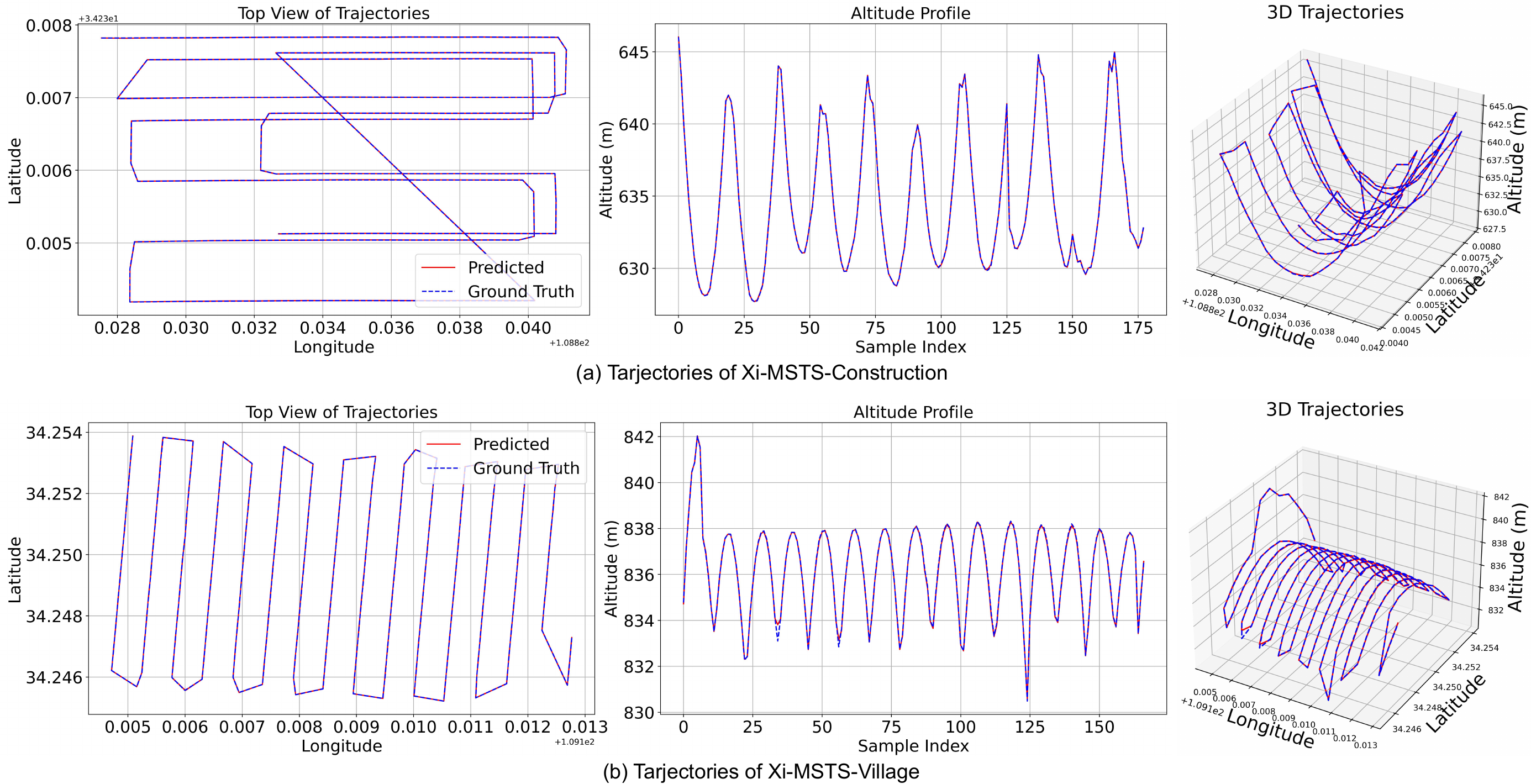}
	\caption{3D trajectory comparisons between our computed poses (red solid lines) and RTK-GPS ground truth (blue dashed lines) for Xi-MSTS-Construction (top) and Xi-MSTS-Village (bottom) scenes, displayed in both top-view and 3D perspectives.}
	\label{fig:traj-vis}
\end{figure*}

\begin{figure*}
	\centering
	\includegraphics[width=.95\textwidth]{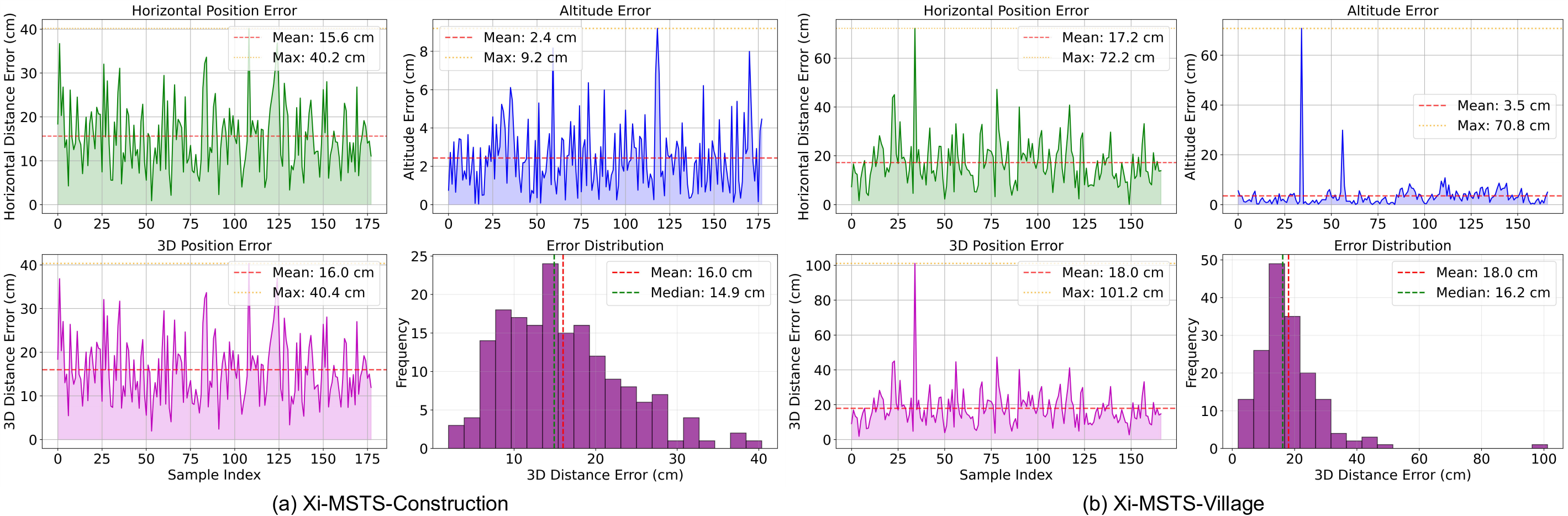}
	\caption{\textbf{Positioning error analysis for Construction and Village.} Comprehensive error analysis showing horizontal position errors (green), altitude errors (blue), 3D position errors (magenta), and error distributions for Xi-MSTS-Construction (left) and Xi-MSTS-Village (right).}
	\label{fig:error_analysis}
\end{figure*}

\subsection{Relocalization Benchmark}

To validate the effectiveness of our proposed method $\mathrm{Hi}^2$-GSLoc, we compare it with state-of-the-art methods on the widely-used Cambridge Landmarks Dataset \cite{kendall2015posenet} for outdoor localization. As shown in Table \ref{tab:cambridge}, we select fourteen representative methods across three categories for comparison: five image-based methods (PoseNet \cite{kendall2015posenet}, MS-Transformer \cite{ms-transformer}, Learn-$\theta^2$ PN \cite{geometric_loss_2017}, LSTM PN \cite{lstm_2016}, and Geo. PN \cite{geometric_loss_2017}), four structure-based methods (SIFT, HSCNet \cite{Li_Wang_Zhao_Verbeek_Kannala_2020}, HLoc \cite{Sarlin_hloc_2019} (SuperPoint \cite{DeTone_SuperPoint_2018} + SuperGlue \cite{Sarlin_superglue_2020}), and DSAC* \cite{brachmann2017dsac}), and five analysis-by-synthesis methods (Dfnet \cite{chen2022dfnet}, NeRFMatch \cite{zhou2024nerfectmatchexploringnerf}, PNeRFLoc \cite{zhao2024pnerfloc}, CROSSFIRE \cite{crossfire_2023}, and GSplatLoc \cite{Sidorov_gsplatloc_2025}). We report the median translation (cm) and rotation errors ($^\circ$) in Table \ref{tab:cambridge}. Previous analysis-by-synthesis methods \cite{zhai2025splatloc, zhao2024pnerfloc, Sidorov_gsplatloc_2025, zhou2024nerfectmatchexploringnerf} have demonstrated superior performance on indoor datasets compared to outdoor scenarios, where structure-based methods typically achieve higher accuracy. Unlike these approaches, our $\mathrm{Hi}^2$-GSLoc maintains competitive performance on outdoor datasets, consistently outperforming structure-based methods in terms of localization precision. Specifically, our method achieves superior rotation accuracy across all evaluated scenes compared to existing approaches. For translation accuracy, $\mathrm{Hi}^2$-GSLoc demonstrates competitive performance in Hospital, Shop, and Church scenes. When averaged across all scenes, $\mathrm{Hi}^2$-GSLoc surpasses all current state-of-the-art methods in both translation and rotation metrics.

\begin{table*}[!t]
\caption{\textbf{Ablation study on $\mathbf{\mathrm{Hi}^2}$-GSLoc pipeline.} Average median errors, recall rates, train and inference time are reported on Mill 19-Rubble and Xi-MSTS with SfM ground truth.}
\label{tab:pipeline-ablation}
    \begin{tabular*}{\textwidth}{l|l|ll|llll|ll}
    \toprule
     & method & AE$\downarrow$ & TE$\downarrow$ & 500/10$^ \circ$$\uparrow$ & 200/5$^ \circ$ $\uparrow$& 5/5$^ \circ$ $\uparrow$& 2/2$^ \circ$ $\uparrow$ & Train/s $\downarrow$ & Inference/s $\downarrow$ \\ \midrule
     & 10000 (initial) & 0.1289 & 0.6425 & 86.73 & 85.71 &  82.65 & 82.14 & 39m02s & 1.1603\\ 
     & 10000 (refine) & 0.0194 & 0.1343 & 87.24 & 87.24  &87.24 & 87.24 &44m02s & 1.2034 \\ 
     & \cellcolor{lightgray}10000 (final) & \cellcolor{lightgray}0.0174 &  \cellcolor{lightgray}0.1206 & \cellcolor{lightgray}100.0 & \cellcolor{lightgray}100.0  &\cellcolor{lightgray}100.0 & \cellcolor{lightgray}100.0 &\cellcolor{lightgray}39m02s &\cellcolor{lightgray}0.00025\\ 
     & 20000 (initial) & 0.1037 & 0.5183 & \textcolor{blue}{91.32} &  90.30 &88.77 &87.75  &77m53s & 1.1508\\ 
     & 20000 (refine) & \textcolor{blue}{0.0136} & \textcolor{blue}{0.1060} & \textcolor{blue}{91.32} & \textcolor{blue}{91.32}   &\textcolor{blue}{91.32} & \textcolor{blue}{91.32} &83m53s & 1.2612\\ 
     & \cellcolor{lightgray}20000 (final) & \cellcolor{lightgray}0.0130 &\cellcolor{lightgray}0.1013 &\cellcolor{lightgray}100.0 & \cellcolor{lightgray}100.0 &  \cellcolor{lightgray}100.0 & \cellcolor{lightgray}100.0 & \cellcolor{lightgray}83m53s &\cellcolor{lightgray}0.00026\\ 
     & 30000 (initial) & 0.0987 & 0.5242 & 93.36 & 91.83 & 90.36 & 88.26 &116m19s & 1.1533\\ 
     & 30000 (refine) & \textcolor{red}{0.0128} & \textcolor{red}{0.1021} & \textcolor{red}{93.87} & \textcolor{red}{93.87} & \textcolor{red}{93.87} & \textcolor{red}{93.87} & 132m39s& 1.2996\\ 
     \multirow{-9}{*}{Mill 19-Rubble} &\cellcolor{lightgray}30000 (final) & \cellcolor{lightgray}{0.0119} & \cellcolor{lightgray}{0.0997} & \cellcolor{lightgray}{100.0} & \cellcolor{lightgray}{100.0} &  \cellcolor{lightgray}{100.0} & \cellcolor{lightgray}{100.0} & \cellcolor{lightgray}132m39s& \cellcolor{lightgray}0.00024\\ 
     \midrule 
     & 10000 (initial) & 0.2032 & 0.6933 & 96.38 & 96.38 &  96.38 & 96.38 & 25m35s & 0.3053 \\ 
     & 10000 (refine) &  0.1334 &  0.5967 & 96.99 &  \textcolor{blue}{96.99} &  \textcolor{blue}96.99  &  \textcolor{blue}96.99 &31m35s & 0.7135 \\ 
     & \cellcolor{lightgray}10000 (final) & \cellcolor{lightgray}0.1319 &\cellcolor{lightgray}0.5621  & \cellcolor{lightgray}100.0 &\cellcolor{lightgray}100.0 & \cellcolor{lightgray}100.0 & \cellcolor{lightgray}100.0 &\cellcolor{lightgray}31m35s & \cellcolor{lightgray}0.00028\\ 
     & 20000 (initial) & 0.1973 & 0.6631  & \textcolor{blue}{98.19} & \textcolor{red}{98.19} &  \textcolor{red}{98.19} &\textcolor{red}{98.19} &53m21s & 0.2882\\ 
     & 20000 (refine) & \textcolor{blue}{0.1111} & \textcolor{blue}{0.5199} & \textcolor{red}{98.79} & \textcolor{red}{98.19} &  \textcolor{red}{98.19} &\textcolor{red}{98.19} & 59m31s & 0.8149\\ 
     &\cellcolor{lightgray}20000 (final) &\cellcolor{lightgray}0.1107 & \cellcolor{lightgray}0.5082 &\cellcolor{lightgray}\cellcolor{lightgray}100.0 & \cellcolor{lightgray}100.0 &   \cellcolor{lightgray}100.0 &\cellcolor{lightgray}100.0 &\cellcolor{lightgray}59m31s& \cellcolor{lightgray}0.00024 \\ 
     & 30000 (initial) & 0.1957 & 0.6612 & \textcolor{blue}{98.19} & \textcolor{red}{98.19} &  \textcolor{red}{98.19} &\textcolor{red}{98.19} & 80m40s& 0.2502 \\ 
     & 30000 (refine) & \textcolor{red}{0.1062} & \textcolor{red}{0.4552} & \textcolor{blue}{98.19} & \textcolor{red}{98.19} &  \textcolor{red}{98.19} &\textcolor{red}{98.19} & 86m53s & 0.8407\\  
     \multirow{-9}{*}{Hills-UE4} & \cellcolor{lightgray}30000 (final) &  \cellcolor{lightgray}\cellcolor{lightgray}{0.1050} & \cellcolor{lightgray}{0.4574} &\cellcolor{lightgray}{100.0} & \cellcolor{lightgray}{100.0} &  \cellcolor{lightgray}{100.0} & \cellcolor{lightgray}{100.0}  & \cellcolor{lightgray}86m53s &\cellcolor{lightgray}0.00025 \\ \midrule
     & 10000 (initial) & 0.1148 & 0.5139 & \textcolor{red}{100.0} &   \textcolor{red}{100.0} & \textcolor{red}{100.0} & 98.31 & 41m51s & 0.7673\\ 
     & 10000 (refine) & \textcolor{red}{0.0259}  &  \textcolor{red}{0.1381} & \textcolor{red}{100.0} & \textcolor{red}{100.0} &   \textcolor{red}{100.0} & \textcolor{red}{100.0}  & 48m58s & 3.8986\\ 
     & \cellcolor{lightgray}10000 (final) &  \cellcolor{lightgray}0.0259 &  \cellcolor{lightgray}0.1381 & \cellcolor{lightgray}100.0 & \cellcolor{lightgray}100.0 &   \cellcolor{lightgray}100.0 & \cellcolor{lightgray}100.0&\cellcolor{lightgray}48m58s &\cellcolor{lightgray}0.00033 \\ 
     & 20000 (initial) & 0.0957  & 0.4434 & \textcolor{red}{100.0} &   \textcolor{red}{100.0} & \textcolor{red}{100.0} & \textcolor{blue}{99.43} & 82m58s & 0.5872  \\ 
     & 20000 (refine) & \textcolor{blue}{0.0276} & \textcolor{blue}{0.1389} & \textcolor{red}{100.0} & \textcolor{red}{100.0} &   \textcolor{red}{100.0} & \textcolor{red}{100.0}  & 90m01s & 4.6024\\ 
     & \cellcolor{lightgray}20000 (final) & \cellcolor{lightgray}0.0276 & \cellcolor{lightgray}0.1389 & \cellcolor{lightgray}100.0 &\cellcolor{lightgray}100.0 &  \cellcolor{lightgray}100.0 & \cellcolor{lightgray}100.0  & \cellcolor{lightgray}90m01s & \cellcolor{lightgray}{0.00032}\\ 
     & 30000 (initial) &{0.0821} & {0.4288} & \textcolor{red}{100.0} & \textcolor{red}{100.0} &   \textcolor{red}{100.0} & \textcolor{red}{100.0} & 123m28s & 0.5506 \\ 
     & 30000 (refine) & {0.0291} & {0.1456} & \textcolor{red}{100.0} & \textcolor{red}{100.0} &   \textcolor{red}{100.0} & \textcolor{red}{100.0} & 131m28s & 4.6657\\ 
     \multirow{-9}{*}{Construction}  & \cellcolor{lightgray}30000 (final) & \cellcolor{lightgray}{0.0291} & \cellcolor{lightgray}{0.1456} & \cellcolor{lightgray}{100.0} & \cellcolor{lightgray}{100.0} &\cellcolor{lightgray}{100.0} & \cellcolor{lightgray}{100.0} & \cellcolor{lightgray}131m28s &\cellcolor{lightgray}0.00028 \\ \midrule
     & 10000 (initial) &  0.1832 &  0.7072 & 96.73 & 96.73 &  96.73 & 96.07 & 24m23s & 0.0976 \\ 
     & 10000 (refine) & 0.0633 & 0.2860 & 97.38 & 97.38 &  97.38 & \textcolor{blue}{97.38} & 28m33s & 1.3850\\ 
     & \cellcolor{lightgray}10000 (final) &  \cellcolor{lightgray}0.0605 &  \cellcolor{lightgray}0.2836 & \cellcolor{lightgray}100.0 &  \cellcolor{lightgray}100.0 &   \cellcolor{lightgray}100.0 & \cellcolor{lightgray}100.0 & \cellcolor{lightgray}28m33s &\cellcolor{lightgray}0.00024 \\ 
     & 20000 (initial) & 0.1500 & 0.6396 &  \textcolor{blue}{98.03} &  \textcolor{blue}{98.03} &   \textcolor{blue}{98.03} & \textcolor{blue}{97.38} & 54m38s & 0.1043\\ 
     & 20000 (refine) & \textcolor{blue}{0.0519} & \textcolor{blue}{0.2229} & \textcolor{red}{98.69} &  \textcolor{red}{98.69}  & \textcolor{red}{98.69} & \textcolor{red}{98.69}  & 50m26s &1.4446 \\ 
     & \cellcolor{lightgray}20000 (final) &\cellcolor{lightgray}0.0512 & \cellcolor{lightgray}0.2219 & \cellcolor{lightgray}100.0 & \cellcolor{lightgray}100.0 & \cellcolor{lightgray}100.0 & \cellcolor{lightgray}100.0 & \cellcolor{lightgray}54m38s &\cellcolor{lightgray}0.00026 \\ 
     & 30000 (initial) & 0.1386 & 0.6359 & \textcolor{red}{98.69} & \textcolor{red}{98.69} &  \textcolor{red}{98.69} & \textcolor{blue}{97.38}& 76m09s&0.1265\\ 
     & 30000 (refine) &  \textcolor{red}{0.0377} & \textcolor{red}{0.1578} & \textcolor{red}{98.69} & \textcolor{red}{98.69} &  \textcolor{red}{98.69} & \textcolor{red}{98.69} & 80m39s & 1.4045\\
     \multirow{-9}{*}{Campus} & \cellcolor{lightgray}30000 (final) & \cellcolor{lightgray}0.0362 &\cellcolor{lightgray}0.1552 &\cellcolor{lightgray}100.0 & \cellcolor{lightgray}100.0 & \cellcolor{lightgray}100.0 & \cellcolor{lightgray}100.0 &\cellcolor{lightgray}80m39s &\cellcolor{lightgray}0.00026  \\ 
     \midrule
     & 10000 (initial) &  0.1703 &  0.6013 & 86.51 & 86.51 &  86.51 & 86.51 &40m36S & 1.0731\\ 
     & 10000 (refine) & 0.0557 & 0.1868 & 88.76 & 88.76 &  88.76 & 88.76 & 48m57s&0.5045\\ 
     & \cellcolor{lightgray}10000 (final) &  \cellcolor{lightgray}0.0501 &  \cellcolor{lightgray}0.1709 & \cellcolor{lightgray}100.0 &  \cellcolor{lightgray}100.0 &   \cellcolor{lightgray}100.0 & \cellcolor{lightgray}100.0 & \cellcolor{lightgray}48m57s & \cellcolor{lightgray}0.00024\\ 
     & 20000 (initial) & 0.1402 & 0.4532 &  92.13 & 92.13 &  92.13 & 92.18 & 76m54s & 1.0006\\ 
     & 20000 (refine) & \textcolor{blue}{0.0459} & \textcolor{blue}{0.1691} & \textcolor{red}{94.38} &  \textcolor{red}{94.38} &  \textcolor{red}{94.38} & \textcolor{red}{94.38} & 85m22s & 0.7126\\ 
     & \cellcolor{lightgray}20000 (final) &\cellcolor{lightgray}0.0443 & \cellcolor{lightgray}0.1611 & \cellcolor{lightgray}100.0 & \cellcolor{lightgray}100.0 & \cellcolor{lightgray}100.0 & \cellcolor{lightgray}100.0 & \cellcolor{lightgray}85m22s & \cellcolor{lightgray}0.00025\\ 
     & 30000 (initial) & 0.1382 & 0.4665 & 92.13 & 92.13 &  92.13 & 92.13 & 112m34s & 0.9844\\ 
     & 30000 (refine) &  \textcolor{red}{0.0432} & \textcolor{red}{0.1550} & \textcolor{blue}{93.82} & \textcolor{blue}{93.82} &  \textcolor{blue}{93.82} & \textcolor{blue}{93.82} &120m59s& 0.8321 \\
     \multirow{-9}{*}{Village} & \cellcolor{lightgray}30000 (final) & \cellcolor{lightgray}0.0410 &\cellcolor{lightgray}0.1508 &\cellcolor{lightgray}100.0 & \cellcolor{lightgray}100.0 & \cellcolor{lightgray}100.0 & \cellcolor{lightgray}100.0 &\cellcolor{lightgray}120m59s &\cellcolor{lightgray}0.00026 \\
     \bottomrule
    \end{tabular*}
\end{table*}

\subsection{Relocalization in Remote sensing}
Building upon these promising results in standard outdoor localization (Table \ref{tab:cambridge}), we further investigate the performance of $\mathrm{Hi}^2$-GSLoc against competitive analysis-by-synthesis and structure-based methods in the more challenging remote sensing domain. We conduct extensive experiments across diverse UAV scenarios with varying flight altitudes, illumination conditions, viewing angles, and terrain types. The evaluation encompasses real flight data, public datasets, and synthetic environments. Table \ref{tab:compare} presents the comprehensive evaluation results. ``Ours'' denotes the pose estimation results after sparse matching and iterative dense matching optimization, while ``Ours (final)'' represents the results after our reliability filtering mechanism. Notably, as shown in the gray-shaded regions of Table \ref{tab:compare}, our filtering mechanism (Consistency Verification) successfully eliminates 100\% of unreliable pose estimates, ensuring robust performance in challenging remote sensing scenarios. The red and blue numbers indicate the best and second-best results among our unfiltered estimates and competing methods, respectively. The results demonstrate that beyond filtering unreliable pose estimates, our $\mathrm{Hi}^2$-GSLoc achieves superior recall rates and the lowest translation and rotation errors across all evaluated datasets, while maintaining efficient inference time.

To further validate the accuracy of our relocalization results, we conduct comprehensive trajectory analysis on the Construction and Village scenes. Figure \ref{fig:traj-vis} presents 2D top-view and 3D trajectory comparisons between our estimated poses and RTK-GPS ground truth. The visualizations demonstrate excellent alignment between our computed trajectories (red solid lines) and RTK-GPS references (blue dashed lines) across both scenes. Quantitative analysis reveals exceptional precision with mean absolute errors of $0.00000092^\circ$ latitude and $0.00000104^\circ$ longitude for Construction, and $0.00000119^\circ$ latitude and $0.00000092^\circ$ longitude for Village.

Figure \ref{fig:error_analysis} provides detailed error characterization for both scenarios. The Construction scene exhibits consistent positioning performance throughout the flight sequence, with mean errors of 15.6 cm horizontally and 2.4 cm in altitude. In contrast, the Village scene, captured at higher flight altitude, demonstrates the impact of increased elevation on measurement precision. While occasional altitude variations (up to 70.8 cm) occur due to the elevated flight conditions, horizontal positioning maintains robustness with a mean error of 17.2 cm. The error distribution histograms indicate that most positioning errors fall within acceptable ranges, yielding median 3D errors of 14.9 cm and 16.2 cm respectively. This evaluation demonstrates the effectiveness of our $\mathrm{Hi}^2$-GSLoc method across varying flight conditions and terrain characteristics, establishing its reliability for real-world remote sensing applications.

\begin{table*}
\caption{Ablation study on consistent render-aware sampling strategy and landmark-guided keypoint detector in initial pose estimation.}
\label{tab:abla-initial}
    \begin{tabular*}{.95\textwidth}{l|l|ll|ll|llll}
    \toprule
    & C.R-A.S & SuperPoint & L-G.D & AE $\downarrow$ & TE $\downarrow$ & 500/10$^ \circ$$\uparrow$ &200/5$^ \circ$ $\uparrow$&5/5$^ \circ$ $\uparrow$& 2/2$^ \circ$ $\uparrow$  \\ 
    \midrule
    &           & $\checkmark$ &  & 105.267 &  1050.596& 27.17 & 25.12  & 18.97 & 14.35 \\ 
    & $\checkmark$ & $\checkmark$ & & \textcolor{blue}{0.1676} & \textcolor{blue}{0.9184} & \textcolor{blue}{82.56} & \textcolor{blue}{80.00}  & \textcolor{blue}{74.35} & \textcolor{blue}{71.28} \\ 
    &  &  & $\checkmark$ & 19.328 & 404.723 & 48.97 & 45.91  & 37.75 & 30.10 \\ 
    \multirow{-4}{*}{Rubble}& $\checkmark$&  &$\checkmark$ & \textcolor{red}{0.0987}&\textcolor{red}{0.5242}& \textcolor{red}{93.36} & \textcolor{red}{91.83} & \textcolor{red}{90.36} & \textcolor{red}{88.26} \\
    \midrule
    &           & $\checkmark$ &  & 0.3211 & 1.0398 & 84.33 & 83.73  & 82.53 & 78.91\\ 
    & $\checkmark$ & $\checkmark$ & & \textcolor{blue}{0.2320} & \textcolor{blue}{0.8183} & \textcolor{blue}{93.97} & \textcolor{blue}{93.97}  & \textcolor{blue}{93.97} & \textcolor{blue}{89.75} \\ 
    &  &  & $\checkmark$ &0.3119  & 0.9557 & 92.77 & 92.16  & 92.16 & 86.14 \\ 
    \multirow{-4}{*}{Hills-UE4}& $\checkmark$&  &$\checkmark$ & \textcolor{red}{0.1957}&\textcolor{red}{0.6612}& \textcolor{red}{98.19} & \textcolor{red}{98.19} & \textcolor{red}{98.19} & \textcolor{red}{98.19} \\
    \midrule
    &           & $\checkmark$ &  &  0.2814& 1.3515 & 85.95 & 84.26  & 83.14 & 70.78 \\ 
    & $\checkmark$ & $\checkmark$ & & \textcolor{blue}{0.1836} & \textcolor{blue}{0.8829} & \textcolor{blue}{99.43} & \textcolor{blue}{99.43}  & \textcolor{blue}{99.43} & \textcolor{blue}{97.75} \\ 
    &  &  & $\checkmark$ & 0.3248  & 1.5097 & 84.26 & 83.70  & 82.02 & 64.60\\ 
    \multirow{-4}{*}{Construction}& $\checkmark$&  &$\checkmark$ & \textcolor{red}{0.0821}&\textcolor{red}{0.4288}& \textcolor{red}{100.0} & \textcolor{red}{100.0} & \textcolor{red}{100.0} & \textcolor{red}{100.0} \\
    \midrule
    &           & $\checkmark$ &  & 0.2189 & 0.8284& 96.73& 96.73 &96.73  &91.50\\ 
    & $\checkmark$ & $\checkmark$ &  & \textcolor{blue}{0.2147}& \textcolor{blue}{0.8218} & \textcolor{blue}{97.38}  & \textcolor{blue}{97.38} & \textcolor{blue}{97.38} & \textcolor{blue}{93.46} \\ 
    &  &  & $\checkmark$ & {0.2256} & {0.8529} & {96.73} & 96.73  & {96.73} &\textcolor{blue}{93.46}  \\ 
    \multirow{-4}{*}{Campus}& $\checkmark$&  &$\checkmark$ & \textcolor{red}{0.1386}&\textcolor{red}{0.6359}& \textcolor{red}{98.69} & \textcolor{red}{98.69} & \textcolor{red}{98.69} & \textcolor{red}{97.38} \\
    \midrule
    &           & $\checkmark$ &   & 101.7418 & 1156.8260 & 26.40 & 24.15  & 23.03 & 15.16 \\ 
    & $\checkmark$ & $\checkmark$ &  & \textcolor{blue}{0.4753} & \textcolor{blue}{1.6064} & \textcolor{blue}{62.71} & \textcolor{blue}{62.14}  & \textcolor{blue}{61.01} & \textcolor{blue}{49.47} \\ 
    &  &    & $\checkmark$ & 58.2024 & 225.4948 & 45.45 &  43.18 &41.47  & 32.38 \\ 
    \multirow{-4}{*}{Village}& $\checkmark$&  &$\checkmark$ & \textcolor{red}{0.1382}&\textcolor{red}{0.4665}& \textcolor{red}{92.13} & \textcolor{red}{92.13} & \textcolor{red}{92.13} & \textcolor{red}{92.13} \\
    \bottomrule
    \end{tabular*}
\end{table*}

\begin{figure}
	\centering
	\includegraphics[width=\columnwidth]{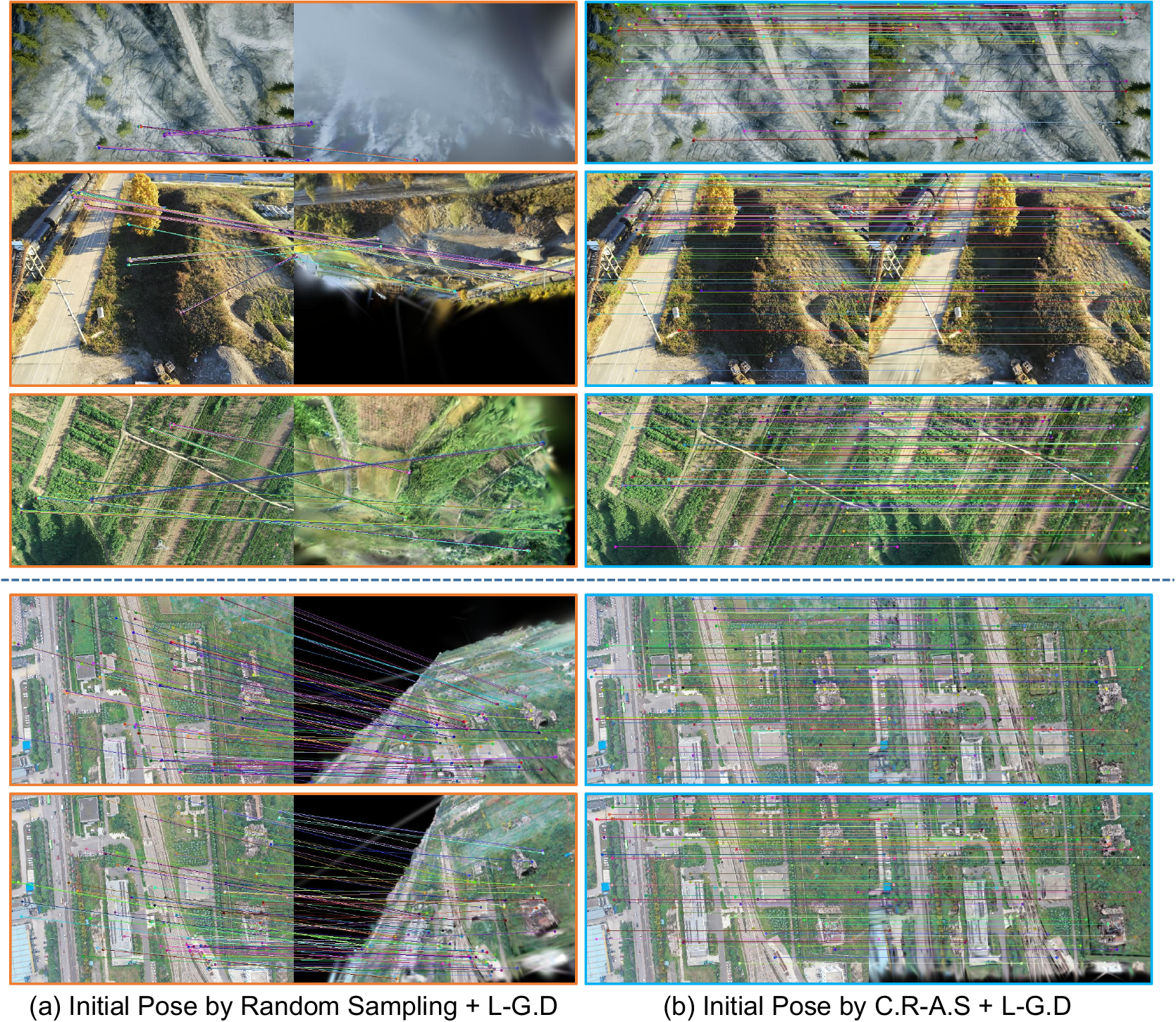}
	\caption{\textbf{Comparison of dense matching results under different initial pose estimation strategies}. (a) Random sampling and (b) our C.R-A.S with L-G.D. Orange boxes highlight incorrect matches from poor rendering, blue boxes show accurate matches. Samples below the blue dashed line have recoverable errors, while those above have excessive errors that cannot be corrected by dense matching.}
	\label{fig:initial-error}
\end{figure}

\subsection{Ablation study}
In this section, we present comprehensive ablation studies to analyze the contribution of each component in our $\mathrm{Hi}^2$-GSLoc framework to relocalization performance.

\textbf{Dual-hierarchical localization pipeline}. In Table \ref{tab:pipeline-ablation}, we report the median errors and recall rates of our algorithm at different stages across five scenes from Xi-MSTS and Mill 19-Rubble datasets. The ``initial'' stage refers to the pose estimation results from sparse matching between retrieved images and Gaussian landmarks. The ``refine'' stage represents the iteratively optimized poses through dense matching with rasterized packages (features, depth, and RGB) rendered from the initial pose. The ``final'' stage denotes the results after consistency checking and filtering of unreliable estimates. The numbers 10,000, 20,000, and 30,000 indicate different training iterations for the scene-specific Gaussian models. The results demonstrate that dense matching consistently improves localization accuracy over the sparse matching stage across all Gaussian model configurations. Our consistency verification mechanism successfully filters out 100\% of unreliable results from any stage of the pipeline while requiring minimal computational overhead (about \textbf{0.24ms} per inference). The majority of scenes achieve optimal performance when using Gaussian models trained for 30,000 iterations, indicating the importance of sufficient training for high-quality scene representation. This ablation study validates the effectiveness and robustness of each component in our hierarchical relocalization pipeline.

\begin{table*}
\caption{Ablation study on dense rasterization matching for pose optimization.}
\label{tab:abla-dense}
    \begin{tabular*}{.95\textwidth}{l|ll|l|ll|llll}
    \toprule
    & SuperPoint & Gaussian & PMM & AE $\downarrow$ & TE $\downarrow$ & 500/10$^ \circ$$\uparrow$ &200/5$^ \circ$ $\uparrow$&5/5$^ \circ$ $\uparrow$& 2/2$^ \circ$ $\uparrow$  \\ 
    \midrule
    & $\checkmark$ &  &  & 116.4793 & 595.9725 & 3.0769 & 1.5384 & 1.0256& 0.5128 \\ 
    & $\checkmark$ &  & $\checkmark$ & 113.3258 & 544.8418 & 4.6153 & 3.0769 & 1.0256& 0.5128 \\ 
    &  & $\checkmark$  &  &  \textcolor{blue}{0.0264}& \textcolor{blue}{0.1242} & \textcolor{blue}{93.36} & \textcolor{blue}{92.82} & \textcolor{blue}{91.76} & \textcolor{blue}{90.77} \\ 
    \multirow{-4}{*}{Rubble}& & $\checkmark$ &$\checkmark$ & \textcolor{red}{0.0128}&\textcolor{red}{0.1021}& \textcolor{red}{93.87} & \textcolor{red}{93.87} & \textcolor{red}{93.87} & \textcolor{red}{93.87} \\
    \midrule
    & $\checkmark$ &  &  & 0.1144 & 0.7321 & 95.78 & 95.78 & 95.78 & 95.78 \\ 
    & $\checkmark$ &  & $\checkmark$ & 0.1076 & 0.7441 & \textcolor{red}{98.19} & \textcolor{red}{98.19} & \textcolor{red}{98.19} & \textcolor{red}{98.19} \\ 
    &  & $\checkmark$  &  & \textcolor{red}{0.1046} & \textcolor{red}{0.4473} & \textcolor{red}{98.19} & \textcolor{red}{98.19} & \textcolor{red}{98.19} & \textcolor{red}{98.19} \\
    \multirow{-4}{*}{Hills-UE4}&  &$\checkmark$ &$\checkmark$ & \textcolor{blue}{0.1062}&\textcolor{blue}{0.4552}& \textcolor{red}{98.19} & \textcolor{red}{98.19} & \textcolor{red}{98.19} & \textcolor{red}{98.19} \\
    \midrule
    & $\checkmark$ &  &  & 122.1261 & 793.8126 & 0.00 & 0.00 & 0.00 &0.00  \\ 
    & $\checkmark$ &  & $\checkmark$ & 130.8981 & 1119.1679 & 0.00 & 0.00 & 0.00&0.00 \\ 
    &  & $\checkmark$  & & \textcolor{blue}{0.0936} & \textcolor{blue}{0.4278} & \textcolor{red}{100.0} & \textcolor{red}{100.0} & \textcolor{red}{100.0} & \textcolor{red}{100.0}\\
    \multirow{-4}{*}{Construction}& & $\checkmark$ &$\checkmark$ & \textcolor{red}{0.0291}&\textcolor{red}{0.1456}& \textcolor{red}{100.0} & \textcolor{red}{100.0} & \textcolor{red}{100.0} & \textcolor{red}{100.0} \\
    \midrule
    & $\checkmark$ &  &  & 0.0899 & 1.0922 & 97.03 & 97.03 & 96.34& 95.03 \\ 
    & $\checkmark$  &  & $\checkmark$ & {0.2256} & {0.8529} & {96.73} & 96.73  & {96.73} &{93.46}  \\ 
    &  & $\checkmark$  & & \textcolor{blue}{0.0682} & \textcolor{blue}{0.2535} & \textcolor{red}{98.69} & \textcolor{red}{98.69} & \textcolor{red}{98.69} & \textcolor{red}{98.69}  \\
    \multirow{-4}{*}{Campus}& & $\checkmark$ &$\checkmark$  & \textcolor{red}{0.0377}&\textcolor{red}{0.1578}& \textcolor{red}{98.69} & \textcolor{red}{98.69} & \textcolor{red}{98.69} & \textcolor{red}{98.69} \\
    \midrule
    & $\checkmark$ &  &  & 17.2569 & 75.0487 & 45.4545 & 43.75 & 42.04& 29.54 \\ 
    & $\checkmark$ &  & $\checkmark$ & 0.7102  & 3.0643 & 56.25 & 53.41 & 51.70 & 42.04\\ 
    &  & $\checkmark$  &  & \textcolor{blue}{0.0526} & \textcolor{blue}{0.1663} & \textcolor{blue}{92.61} & \textcolor{blue}{92.61} & \textcolor{blue}{92.61} & \textcolor{blue}{92.61} \\
    \multirow{-4}{*}{Village}& & $\checkmark$ &$\checkmark$ &  \textcolor{red}{0.0432}&\textcolor{red}{0.1550}& \textcolor{red}{93.82} & \textcolor{red}{93.82} & \textcolor{red}{93.82} & \textcolor{red}{93.82} \\
    \bottomrule
    \end{tabular*}
\end{table*}

\begin{figure*}
	\centering
	\includegraphics[width=\textwidth]{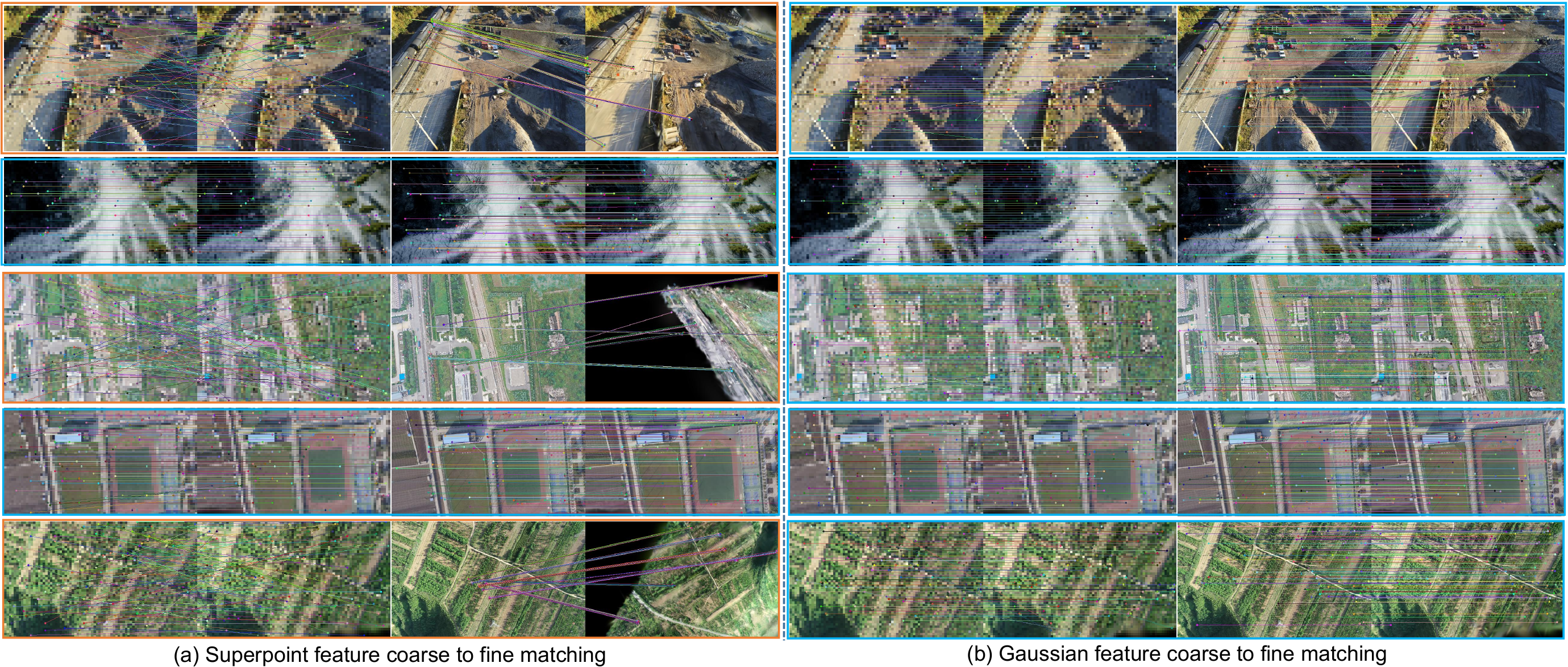}
	\caption{\textbf{Comparison of coarse to fine dense matching with different dense feature extractor.} (a) SuperPoint-based features and (b) our Gaussian features.  From left to right in both (a) and (b): initial coarse matching and iterative dense matching.}
	\label{fig:coarse-fine}
\end{figure*}

\textbf{Initial pose estimation.} The accuracy of initial pose estimation directly determines the quality of rendered images, which subsequently affects dense matching precision. We conduct ablation studies on five scenes from the Mill 19-Rubble and Xi-MSTS datasets to evaluate the effectiveness of our C.R-A.S and L-G.D components, as shown in Table \ref{tab:abla-initial}.
For each scene, we compare four configurations: random sampling strategy, SuperPoint detector, and our proposed C.R-A.S for landmark selection combined with L-G.D for query image keypoint detection. The results demonstrate that using both C.R-A.S and L-G.D consistently achieves the lowest pose errors and highest recall rates across all scenes. Notably, in Mill 19-Rubble and Xi-MSTS-Village, the combination of C.R-A.S and L-G.D achieves 73.91\% and 76.97\% higher recall rates respectively compared to the random sampling and SuperPoint baseline, demonstrating that our modules significantly improve initial pose accuracy while providing greater stability.

Figure \ref{fig:initial-error} visualizes the render and dense matching results under different initial pose estimation strategies. Subfigure (a) shows results obtained using random sampling for initial pose estimation, while subfigure (b) presents results from our proposed Consistent Render-Aware Sampling strategy. Orange boxes highlight dense matching results under incorrectly rendered views, whereas blue boxes indicate accurate matching results under precise initial pose rendering. Additionally, samples below the blue dashed line represent dense matching results with relatively small initial pose errors, which can be further refined through subsequent iterative dense matching optimization. In contrast, samples above the dashed line suffer from excessive initial pose errors, and even dense matching cannot recover accurate poses from such poor initialization. This confirms that accurate initial pose estimation is a critical prerequisite for reliable system localization.

\textbf{Dense rasterization matching.} We compare feature extraction strategies (rendered Gaussian features vs. extracting features from rendered RGB images using existing feature extraction networks) and matching strategies (with/without probabilistic mutual matching (PMM)). Results are presented in Table \ref{tab:abla-dense}. The experimental results reveal significant performance degradation when employing SuperPoint for feature extraction across multiple scenes. Notably, in the Rubble, Construction, and Village scenes, all evaluation metrics demonstrate substantial decline, with the Construction scene experiencing complete localization failure (all recall metrics drop to 0\%).  This performance collapse can be attributed to SuperPoint's limited generalization capability on high-resolution aerial imagery, which is characterized by repetitive patterns and extreme viewpoint variations typical of remote sensing scenarios.

Figure \ref{fig:coarse-fine} provides a qualitative comparison between initial coarse matching and iterative dense matching results, contrasting SuperPoint-based features (left) with our trained Gaussian features (right). The visualization clearly demonstrates that inappropriate feature extraction methods lead to erroneous dense pose estimation, even when accurate initial views are rendered from correct initial poses. This error propagation results in progressively deteriorating views and poses through iterations, creating an unrecoverable optimization failure. These results demonstrate that our Gaussian-based scene-specific feature extraction approach achieves superior robustness and reliability compared to generic feature descriptors. The performance degradation of pre-trained models like SuperPoint can be attributed to the significant domain gap between existing training datasets and remote sensing imagery. 

\begin{figure}
	\centering
	\includegraphics[width=.9\columnwidth]{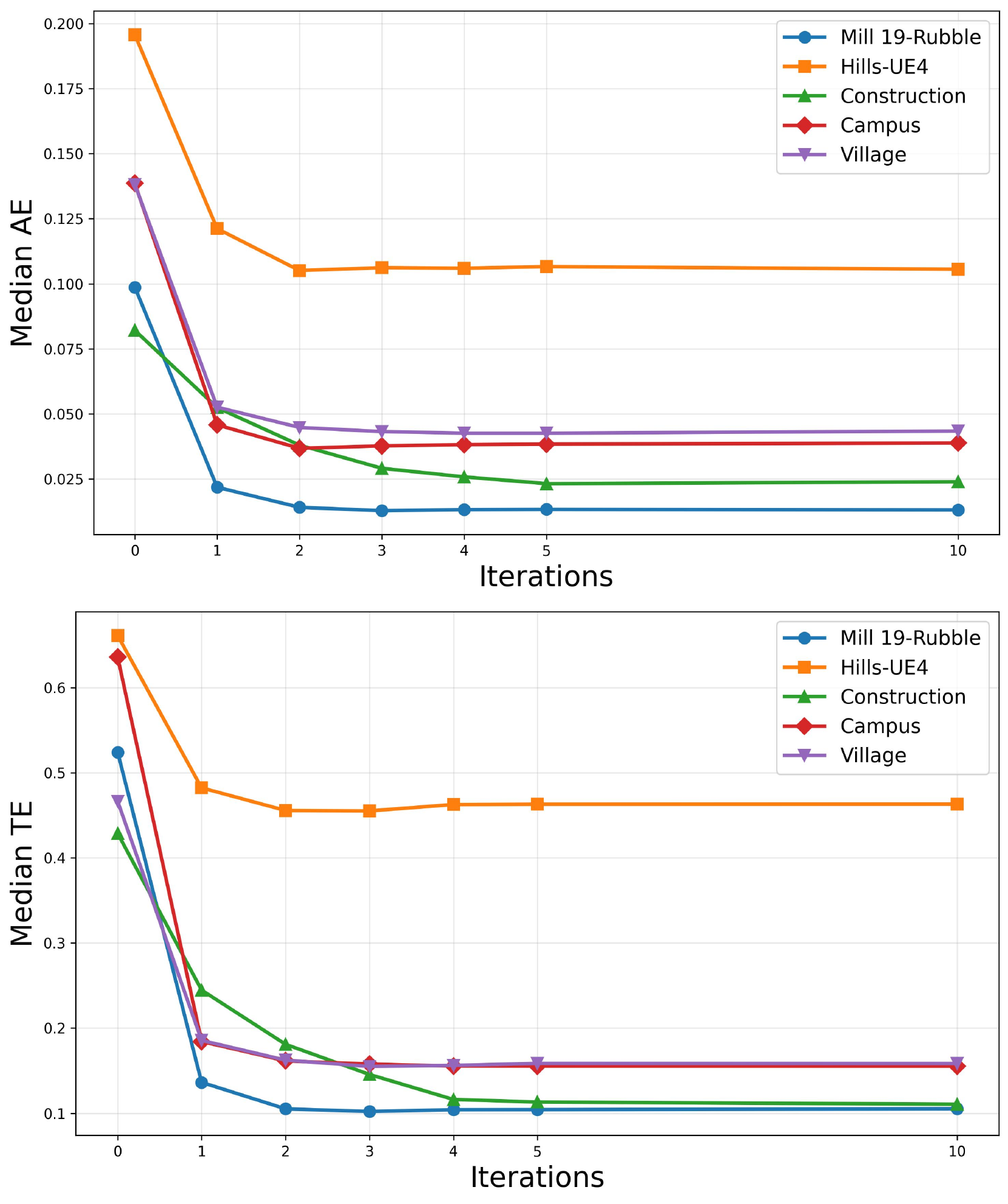}
	\caption{the Median AE (top) and Median TE (bottom) for five datasets over iterations 0-10.}
	\label{fig:iter}
\end{figure}

Additionally, to determine the optimal number of iterations for our method, we conducted an ablation study across five diverse datasets. As illustrated in Figure \ref{fig:iter}, both the median AE and median TE demonstrate rapid convergence within the first few iterations. The results show that most datasets achieve significant error reduction by iteration 3, with marginal improvements observed in subsequent iterations. Specifically, the angular error stabilizes around iteration 3 across all datasets, while the translation error exhibits similar convergence behavior. Considering the trade-off between accuracy and computational efficiency, we selected iteration 3 as the optimal configuration for our method. This choice ensures that our approach maintains high localization accuracy while keeping the inference time reasonable for practical applications. 

\FloatBarrier

\section{Conclusion}

This paper introduces 3D Gaussian Splatting (3DGS) as a novel map representation for visual relocalization, significantly expanding the capabilities of UAV navigation in large-scale remote sensing scenarios. Building upon this scene representation, we present $\mathrm{Hi}^2$-GSLoc, a dual-hierarchical relocalization framework that systematically addresses critical challenges in remote sensing through three key technical innovations: (1) Scalable scene processing through partitioned Gaussian training coupled with dynamic memory management, enabling efficient handling of large-scale environments; (2) Scene-specific feature learning via consistent render-aware landmark sampling that effectively exploits Gaussian geometric constraints to enhance feature representation quality; (3) Robust and accurate pose estimation through a coarse-to-fine refinement strategy with consistency validation, ensuring reliable localization under challenging conditions. Comprehensive experimental evaluation on the Mill 19-Rubble and Xi-MSTS datasets demonstrates the effectiveness and practical utility of our approach. $\mathrm{Hi}^2$-GSLoc achieves superior recall rates, maintains computational efficiency, and delivers centimeter-level accuracy at high altitudes using only visual information. Furthermore, the method exhibits exceptional robustness by effectively filtering unreliable localizations through our consistency validation mechanism, making it particularly well-suited for practical UAV applications in demanding remote sensing environments.


\printcredits
\section*{Declaration of competing interest}

The authors declare that they have no known competing financial interests or personal relationships that could have appeared to influence the work reported in this paper.

\section*{Acknowledgments}

This work was supported by the National Natural Science Foundation of China (NSFC) under Grant No. 42130112 and the Postdoctoral Fellowship Program of CPSF under Grant No. GZB20240986.

\bibliographystyle{cas-model2-names}
\bibliography{cas-refs}

\end{document}